\ifdefined\pdfobjcompresslevel\pdfobjcompresslevel=0\fi
\RequirePackage{fix-cm}
\documentclass{article} 
\usepackage{iclr2027_conference,times}

\usepackage{amsmath,amsfonts,bm}

\def\eqref#1{equation~\ref{#1}}

\def\1{\bm{1}}

\DeclareMathAlphabet{\mathsfit}{\encodingdefault}{\sfdefault}{m}{sl}
\SetMathAlphabet{\mathsfit}{bold}{\encodingdefault}{\sfdefault}{bx}{n}

\usepackage{subcaption}
\usepackage{booktabs}
\usepackage{adjustbox}
\usepackage{multirow}
\usepackage{algorithm}
\usepackage{algpseudocode}

\usepackage{hyperref}
\usepackage{graphicx}
\usepackage{url}
\title{Self-Aligned Forcing: Streaming Video\\
Diffusion with Differentiable Noisy History}

\author{
\textbf{Weiqiang Wang}\textsuperscript{\normalfont 1}\thanks{Equal contribution.} \quad
\textbf{Zhuokun Chen}\textsuperscript{\normalfont 1}\footnotemark[1] \quad
Yusheng Dai\textsuperscript{\normalfont 1} \quad
Boying Li\textsuperscript{\normalfont 1} \quad
Yi Zhang\textsuperscript{\normalfont 2}\thanks{Corresponding authors.} \quad \\
Hossein Rahmani\textsuperscript{\normalfont 3} \quad
Qiuhong Ke\textsuperscript{\normalfont 1}\footnotemark[2] \quad
Jianfei Cai\textsuperscript{\normalfont 1} \\
\textsuperscript{1}Monash University \quad
\textsuperscript{2}Vivix AI \quad
\textsuperscript{3}Lancaster University
}

\usepackage{xspace}
\newcommand{\methodname}{SAF\xspace}
\iclrpreprint
\begin{document}

\maketitle

\begin{abstract}
Autoregressive video diffusion enables interactive streaming generation, but suffers from error accumulation over long rollouts.
Self-rollout training reduces exposure bias, yet finite rollouts leave long-range drift unresolved.
We observe that the noise level of the history key-value (K/V) representations trades visual quality against motion, and that restoring gradients through the history aligns causal training far more closely with bidirectional training.
Motivated by these observations, we introduce \textbf{Self-Aligned Forcing} (\textbf{\methodname}), a training scheme that \textit{aligns} the history of each block with the noise level of the block being denoised.
Specifically, the history is the K/V produced by preceding blocks at the same denoising stage, so all blocks at a stage can be denoised in a single forward pass under a causal mask.
This keeps the noisy history differentiable, allowing future losses to optimize how it is encoded.
\methodname therefore avoids a separate no-gradient rollout and per-block timestep-zero recaching, training up to $1.8\times$ faster than prior methods with lower memory.
At inference, \methodname achieves the highest single-GPU throughput among existing methods and keeps one history bank per stage for a multi-GPU pipeline, reaching $49.1$~FPS on 4 GPUs.
Experiments show superior long-horizon generation with a better balance between visual quality and motion. Project page: \url{https://anonymous.4open.science/w/self-aligned-forcing/}.
\end{abstract}

\section{Introduction}

\begin{figure*}[h]
    \centering
    \includegraphics[width=0.99\textwidth]{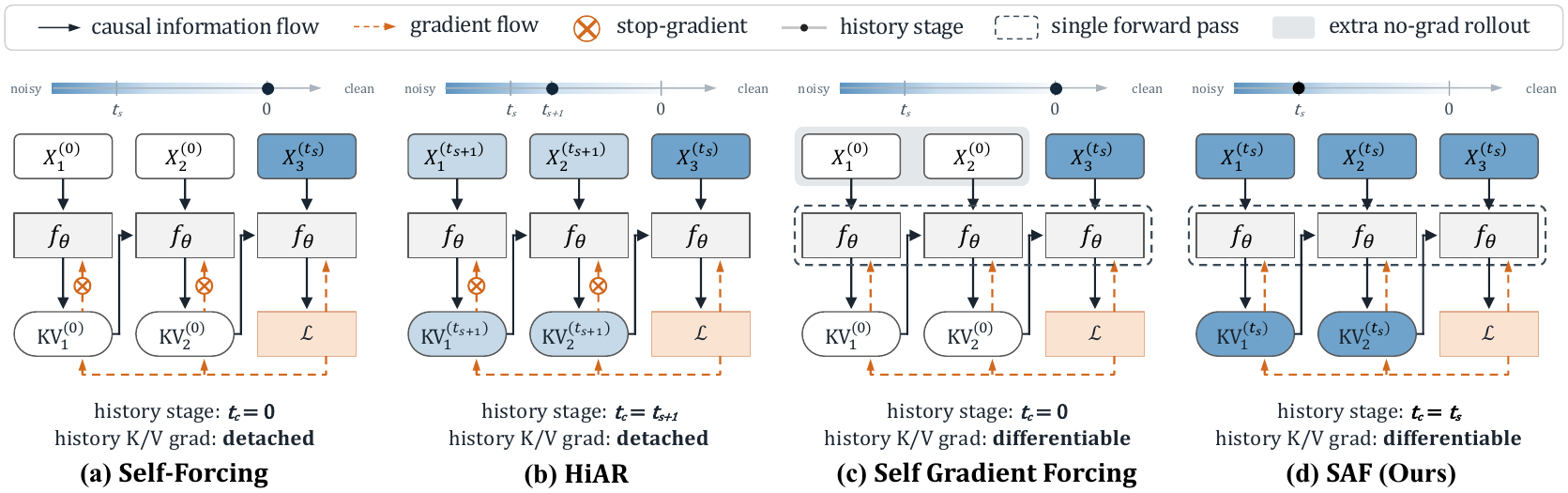}
    \caption{
    \textbf{History stage and gradient flow.}
    The model $f_\theta$ generates blocks from left to right, and the K/V of earlier blocks form the history of the last block, whose loss $\mathcal L$ is shown.
    Superscripts give noise levels ($t_s$: current block, $t_c$: history, $0$: clean).
    Self-Forcing \textbf{(a)} and HiAR \textbf{(b)} detach the history, so the loss cannot reach it; Self Gradient Forcing \textbf{(c)} and \methodname \textbf{(d)} keep this gradient with the history at $t_c=0$ and $t_c=t_s$, respectively, but the former needs an extra no-grad rollout.
    }
    \label{fig:forcing_comparison}
\end{figure*}

Autoregressive (AR) video diffusion has emerged as a promising approach to interactive streaming video generation~\citep{yin2025slow,huang2026self,yang2025longlive}.
It generates a video block by block, conditioning each new block on the key-value (K/V) representations of previously generated blocks, which form the \emph{history}.
The model can thus continuously extend a video and respond to new instructions without reprocessing the full sequence, as required by interactive applications such as game engines and world models~\citep{bruce2024genie,valevski2025diffusion,he2025matrixgame2}.
However, long-horizon streaming remains challenging, as it requires stability over videos much longer than the training clips, meaningful motion, and efficient generation.

Long-horizon stability is limited mainly by error accumulation.
Self-Forcing~\citep{huang2026self} narrows the train--test gap, known as exposure bias~\citep{ranzato2016sequence}, by training the model on its own generated history (Figure~\ref{fig:forcing_comparison}a).
Nevertheless, its finite rollouts do not fully account for errors accumulated over longer horizons.
To improve robustness under limited training rollouts, recent methods build the history from partially denoised states.
Rolling Forcing~\citep{liu2025rolling} jointly denoises a rolling window with staggered noise levels, while HiAR~\citep{zou2026hiar} takes the history one denoising step cleaner than the current block (Figure~\ref{fig:forcing_comparison}b).
HiAR further finds that aligning the history noise level with that of the current block minimizes error accumulation but lowers the overall VBench quality score~\citep{huang2024vbench}.

\begin{figure*}[t]
    \centering
    \begin{minipage}{0.98\textwidth}
    \captionsetup[subfigure]{skip=1pt}
    \subcaptionbox{Noisier history adds motion but lowers quality.
        \label{fig:motivation_stage}}[0.49\linewidth]{%
        \includegraphics[width=0.88\linewidth]{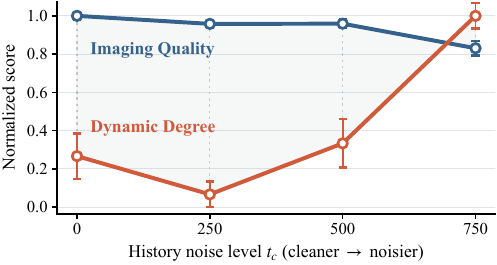}}
    \hfill
    \subcaptionbox{History gradients align with bidirectional training.
        \label{fig:motivation_optimization}}[0.49\linewidth]{%
        \includegraphics[width=0.88\linewidth,trim=0 7bp 0 3bp,clip]{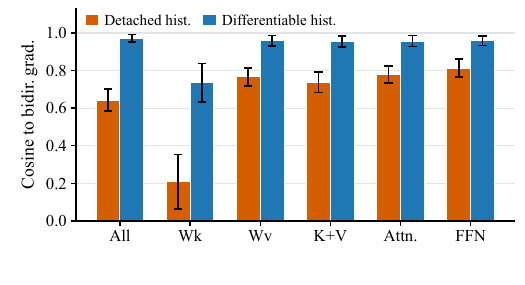}}
    \end{minipage}
    \par\vspace{-6pt}
    \captionsetup{skip=0pt}
    \caption{
    \textbf{Two observations motivating differentiable noisy history.}
    \textbf{(a)} A fixed LongLive model with varying history noise level $t_c$; imaging quality and dynamic degree~\citep{huang2024vbench} are normalized to their values at $t_c=0$ and $t_c=750$, respectively.
    \textbf{(b)} With all settings identical except whether gradients flow through the history K/V, differentiable history raises the gradient cosine similarity to a matched-weight bidirectional reference from $0.643$ to $0.972$.
    Error bars show the range over multiple samples.
    }
    \label{fig:motivation}
\end{figure*}

\textbf{Observation 1.} \textbf{Noisy history not only reduces error accumulation but also increases motion dynamics, at the cost of imaging quality.}
Using a LongLive checkpoint trained with clean history, we vary only the history noise level $t_c$ at inference.
As shown in Figure~\ref{fig:motivation_stage}, sufficiently noisy history substantially increases the VBench~\citep{huang2024vbench} dynamic degree, but imaging quality drops.
The motion gain is notable, since models trained with distribution matching distillation (DMD)~\citep{yin2024improved,yin2024onestep} commonly produce static outputs due to its mode-seeking objective~\citep{zou2026hiar,you2026adaptive,wu2026sgmd}.
HiAR attributes reduced error accumulation to the lower signal-to-noise ratio (SNR) of noisy history~\citep{zou2026hiar}; our observation suggests that this mechanism also underlies the motion gain and the quality loss.
By weakening the constraint that the history imposes on the current block, lower SNR suppresses error propagation and increases motion dynamics, but also reduces temporal consistency and lowers imaging quality.
The benefit and the cost thus share a single cause, yet the model cannot learn to balance them.

\textbf{Observation 2.} \textbf{Detaching the history prevents the model from learning how to encode it for future blocks.}
Existing self-rollout methods detach the history K/V from gradient computation (Figure~\ref{fig:forcing_comparison}a,b), as backpropagating through it would nest the computation graphs of all preceding blocks.
Future losses therefore cannot optimize how preceding blocks are encoded for subsequent prediction.
To quantify this effect, we compare the resulting parameter gradient with a matched-weight bidirectional reference, in which all blocks are processed jointly so that losses on later blocks also supervise earlier ones.
With all other settings identical, restoring the gradient path through the history increases their cosine similarity from $0.643$ to $0.972$ (Figure~\ref{fig:motivation_optimization}).
The gap is largest for the key projection $W_k$, whose cosine similarity drops to about $0.2$ under detachment, indicating that the model barely learns to encode the history for future retrieval.
Noisy history should therefore remain differentiable rather than serve only as fixed conditioning.

Existing methods face a dilemma: noisy history is robust but detached, whereas differentiable history requires costly reconstruction.
Self Gradient Forcing (SGF)~\citep{zhuang2026self} restores gradients through the history (Figure~\ref{fig:forcing_comparison}c), but requires a separate no-gradient rollout to construct the history (similar to SDF in Figure~\ref{fig:sdf_framework}).
Moreover, its history is built from clean states, available only after a block is fully denoised and recached at timestep zero.
This recaching adds a forward pass per block and prevents pipelining across denoising stages, limiting both training and inference efficiency.

Our key insight is that stage alignment between the history and the current block makes noisy history differentiable efficiently without reconstruction.
The history K/V is then exactly what preceding blocks compute while being denoised at the same stage.
Since the history and the current block share one noise level, they can be processed in a single forward pass (Figures~\ref{fig:forcing_comparison}d and~\ref{fig:method_overview}b), so the history is both noisy and differentiable without reconstruction. Besides, this causal training strategy shares the structure of bidirectional diffusion training, in which all blocks share one noise level and later blocks supervise earlier ones.
Notably, this aligned setting lowers quality in HiAR, where the history is detached; once the history is learned, it achieves the highest Quality under every inference policy among the history schedules we study in training (Section~\ref{sec:ablations}).

Building on this insight, we propose \textbf{Self-Aligned Forcing} (\methodname).
SAF rolls out all blocks stage by stage: at each denoising step, a single block-causal forward pass processes every block, and gradients are taken at one randomly sampled stage under the standard DMD objective.
Apart from a clean attention sink~\citep{xiao2024efficient} recached within the same forward pass, it needs neither per-block recaching nor a separate no-gradient rollout, reducing training time and memory (Table~\ref{tab:training_efficiency}).
At inference, SAF can keep one history bank per noise level for a multi-GPU pipeline, or a single bank on one GPU to save memory (Figure~\ref{fig:method_overview}c).
Our contributions are threefold:
\begin{itemize}\setlength{\itemsep}{1pt}\setlength{\parskip}{0pt}\setlength{\topsep}{2pt}
    \item We study how the history should be represented and reused in autoregressive video diffusion, showing that propagating gradients through noisy history K/V brings causal training close to bidirectional training, improving both overall quality and motion dynamics.
    \item We propose \textbf{\methodname}, which aligns the history noise level with the current block. All blocks are then rolled out in parallel within each stage, and gradients at a sampled stage can flow through the history. A clean sink anchors appearance at negligible cost, and per-stage history banks enable pipelined inference across GPUs.
    \item We evaluate SAF extensively on multiple benchmarks. It performs best in the chunkwise setting and competitively in the framewise setting, achieving the highest training and inference efficiency in both settings, e.g., $1.8\times$ faster framewise training than SGF and $49.1$~FPS in chunkwise four-GPU pipelined inference.
\end{itemize}

\section{Related Work}

\textbf{Autoregressive Video Diffusion.}
Autoregressive video diffusion generates a video as causal temporal blocks, enabling streaming generation with a KV-cached history~\citep{teng2025magi,lin2025apt2,kodaira2025streamdit,lu2025reward,yi2025deepforcing}.
Diffusion Forcing~\citep{chen2024diffusion} assigns different noise levels to temporal tokens, CausVid~\citep{yin2025slow} distills a bidirectional model into a few-step causal generator, and LongLive~\citep{yang2025longlive} scales this paradigm to real-time long videos with streaming long tuning, windowed attention, frame sinks, and KV recaching.
To improve long-horizon robustness, Rolling Forcing~\citep{liu2025rolling} jointly denoises a window at staggered noise levels, and HiAR~\citep{zou2026hiar} denoises timestep-first and builds the history from intermediate denoising states.
These methods change the noise level of the history but keep it detached, so losses on later blocks cannot optimize how it is encoded.

\textbf{Self-Generated Rollout and Cross-Block Credit Assignment.}
Self-Forcing~\citep{huang2026self} trains the model on its own generated history to narrow the train--test gap, and Resampling Forcing~\citep{guo2025end} and BAgger~\citep{po2025bagger} further improve robustness through self-resampling and corrective rollout trajectories.
All of them, however, detach the history K/V, so future losses cannot shape how earlier blocks are encoded.
Live Avatar~\citep{huang2025liveavatar} also matches the history timestep to the current block, yet its rollout remains serial and its history detached.
Video-Mirai~\citep{yu2026video} adds future-aware supervision through an auxiliary foresight pathway and feature distillation, whereas SGF~\citep{zhuang2026self} restores future-to-history gradients by re-encoding clean history built in a separate no-gradient rollout.
Existing methods thus either detach the history, add auxiliary supervision, or require a separate no-gradient rollout, leaving efficient future-to-history optimization of noisy history K/V unresolved.

\section{Method}
\label{sec:method}

\subsection{Overview}
\label{sec:overview}

\begin{figure*}[t]
    \centering
    \includegraphics[width=0.95\textwidth]{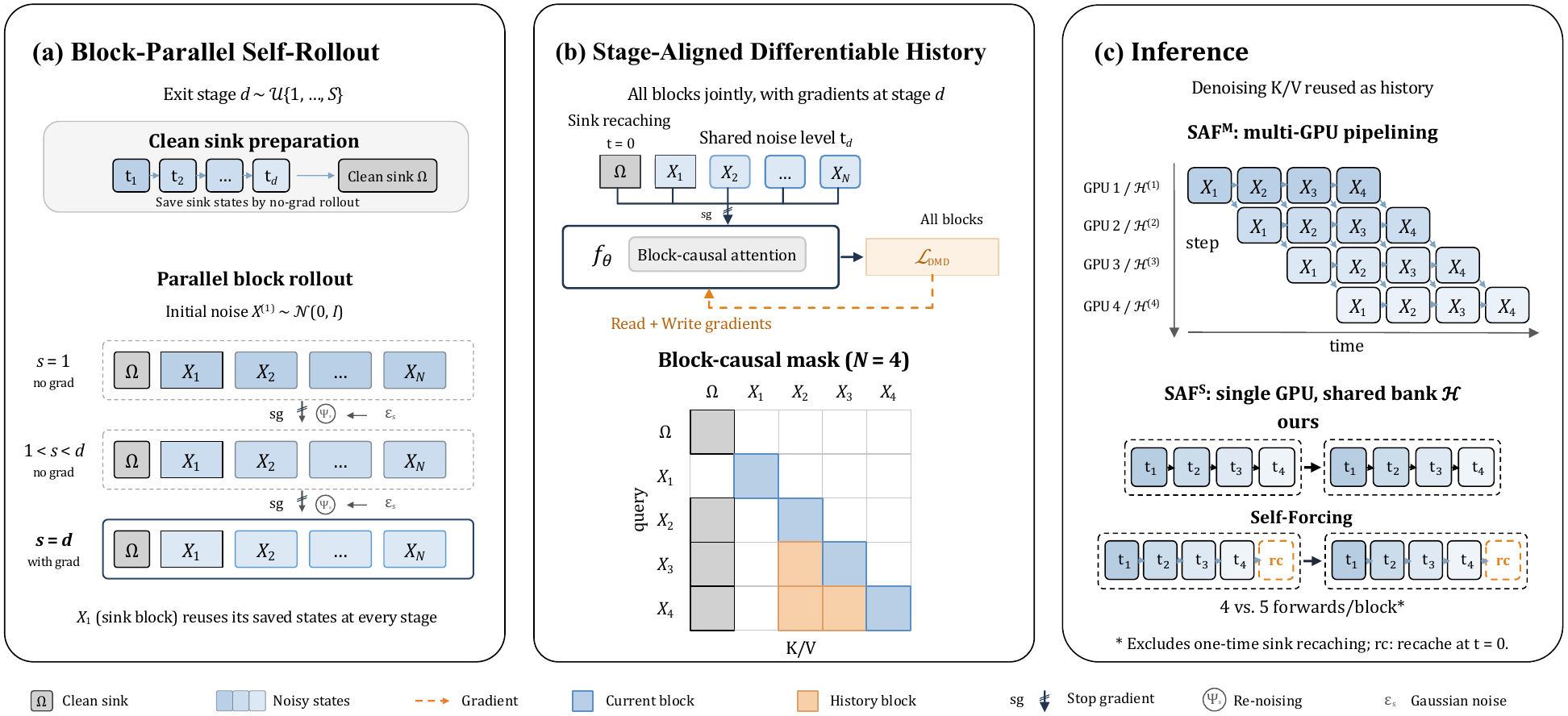}
    \caption{
    \textbf{Overview of \methodname.}
    \textbf{(a)} Rollout is parallel across blocks and sequential only across stages (Section~\ref{sec:parallel_rollout}).
    \textbf{(b)} Stage alignment enables joint denoising with differentiable history in one block-causal forward pass (Section~\ref{sec:aligned_history}).
    \textbf{(c)} Reusing the K/V computed during denoising as history allows multi-GPU stage pipelining and removes per-block recaching (Section~\ref{sec:streaming_inference}).
    }
    \label{fig:method_overview}
\end{figure*}

\methodname rests on a single design choice: at every denoising step, the history is taken at the same noise level as the current block.
This choice makes the local history both noisy and differentiable without any reconstruction (Section~\ref{sec:aligned_history}) and allows self-rollout to proceed in parallel across blocks during training (Section~\ref{sec:parallel_rollout}).
We complement it with a clean sink that anchors appearance over the whole video (Section~\ref{sec:clean_sink}), and at inference it yields per-stage history K/V banks that pipeline naturally across GPUs (Section~\ref{sec:streaming_inference}).
Figure~\ref{fig:method_overview} summarizes the framework.

A video is generated as $N$ blocks, each denoised at noise levels $t_1>\cdots>t_S$, where $t_1=1000$ is pure noise; we call the step at $t_s$ stage~$s$.
At stage $s$, the causal denoiser $f_\theta$ reads the noisy latent $X_i^{(s)}$ of block $i$, its history K/V bank $\mathcal{H}_i^{(s)}$, and the text prompt $p$, predicts a clean-latent estimate $\widehat{X}_i^{(s)}$, and produces the block's own $\mathrm{KV}_i^{(s)}$.
We write this step as $(\widehat{X}_i^{(s)},\mathrm{KV}_i^{(s)})=f_\theta(X_i^{(s)},t_s;\mathcal{H}_i^{(s)},p)$.
$\mathcal{H}_i^{(s)}$ holds the $\mathrm{KV}_{\Omega}$ of a persistent sink $\Omega$ (the first few blocks~\citep{xiao2024efficient}) and the K/V of the latest preceding blocks, which together with block $i$ span a local window of $L$ frames.

\subsection{Stage-Aligned Differentiable History}
\label{sec:aligned_history}

Observations~1 and~2 call for a history that is noisy yet differentiable, but serial self-rollout makes these requirements conflict.
When blocks are generated one after another, the history of block $i$ is produced by earlier forward passes, and keeping it differentiable would nest the computation graphs of all preceding blocks.
Existing methods therefore either detach a noisy history~\citep{zou2026hiar} or construct a clean one in a separate no-gradient rollout~\citep{zhuang2026self}.

Stage alignment resolves this conflict by letting the history be produced and consumed in the same forward pass (Figures~\ref{fig:forcing_comparison}d and~\ref{fig:method_overview}b).
When block $i$ reads its history at its own noise level $t_s$, the K/V of each predecessor $j$ is exactly what block $j$ computes for its own prediction at stage $s$.
A single block-causal forward pass over all blocks therefore yields every prediction together with every history:
\begin{equation}
    \bigl(\widehat{X}^{(s)},\,\mathrm{KV}^{(s)}\bigr)
    =
    f_\theta\bigl(
        \operatorname{sg}(X^{(s)}),
        t_s;
        M_{\mathrm{causal}},
        p
    \bigr),
    \label{eq:aligned_rollout}
\end{equation}
where $X^{(s)}$, $\widehat{X}^{(s)}$, and $\mathrm{KV}^{(s)}$ stack the inputs, estimates, and K/V of all blocks and $\operatorname{sg}$ is stop-gradient.
The block-causal mask $M_{\mathrm{causal}}$ lets each block $i$ attend only to itself and its history $\mathcal{H}_i^{(s)}=\{\mathrm{KV}_j^{(s)}\}_{j\in\mathcal{W}(i)}$, the same-stage K/V of its predecessors within the window $\mathcal{W}(i)$.
Eq.~\ref{eq:aligned_rollout} is thus the per-block step of Section~\ref{sec:overview} applied to all blocks at once, and within a stage, training resembles bidirectional diffusion training with only the future masked out.
For clarity, we omit the sink $\Omega$ here and describe it in Section~\ref{sec:clean_sink}.

Because Eq.~\ref{eq:aligned_rollout} computes the history rather than loading it from a detached cache, the gradient of $\widehat{X}_i^{(s)}$ with respect to the parameters $\theta$ splits into two terms:
\begin{equation}
    \frac{\mathrm{d}\widehat{X}_i^{(s)}}{\mathrm{d}\theta}
    =
    \underbrace{\frac{\partial\widehat{X}_i^{(s)}}{\partial\theta}}_{\text{read}}
    +
    \underbrace{\sum_{h\in\mathcal{H}_i^{(s)}}
    \frac{\partial\widehat{X}_i^{(s)}}{\partial h}\,
    \frac{\mathrm{d}h}{\mathrm{d}\theta}}_{\text{write}},
    \label{eq:history_grad}
\end{equation}
where $h$ ranges over the K/V entries of $\mathcal{H}_i^{(s)}$.
Viewing the history as a memory that predecessors write and block $i$ reads, the read term updates how block $i$ uses the history, while the write term updates how its predecessors encode it, providing future-to-history supervision. As the inputs are detached, the write term changes only how the history is encoded, not what is sampled.
The model thus learns to write noisy history rather than merely tolerate it.

\subsection{Block-Parallel Self-Rollout}
\label{sec:parallel_rollout}

Although each stage of SAF mirrors bidirectional training under a causal mask, the two still differ in the origin of their inputs.
Bidirectional training noises ground-truth videos, which carries over to inference because all frames are denoised jointly.
A causal model, however, reads a history derived from its own earlier outputs at inference, so training on ground-truth inputs creates a train--test mismatch that lets errors accumulate across blocks~\citep{huang2026self}. SAF therefore retains self-rollout: starting from $X^{(1)}\sim\mathcal{N}(0,I)$, the input to each stage is derived from the model's estimate at the preceding stage,
\begin{equation}
    X^{(s+1)}
    =
    \Psi_s\bigl(
        \operatorname{sg}(\widehat{X}^{(s)}),
        \epsilon_s
    \bigr),
    \label{eq:stage_transition}
\end{equation}
where the sampler $\Psi_s$ re-noises the estimate with noise level $t_{s+1}$ using Gaussian noise $\epsilon_s$, and $\widehat{X}^{(s)}$ is given by Eq.~\ref{eq:aligned_rollout}.
In contrast to existing serial self-rollout~\citep{huang2026self,zou2026hiar}, which advances one block per forward pass, SAF parallelizes the rollout across blocks within each stage and keeps it sequential only across stages (Figure~\ref{fig:method_overview}a).

To avoid storing the computation graphs of all $S$ chained stages, we follow the stochastic gradient truncation of Self-Forcing~\citep{huang2026self}: each generator update samples an exit stage $d\sim\mathcal{U}\{1,\ldots,S\}$, covering every noise level in expectation with a single stage in the computation graph.
Stages $s<d$ run without gradients, and only stage $d$ evaluates Eq.~\ref{eq:aligned_rollout} with gradients under the standard DMD objective~\citep{yin2024improved}:
\begin{equation}
    \mathcal{L}(\theta)
    =
    \mathcal{L}_{\mathrm{DMD}}\bigl(\widehat{X}^{(d)},p\bigr),
    \label{eq:saf_objective}
\end{equation}
with score models updated as in DMD (Algorithm~\ref{alg:aligned_saf}).

Block-parallel rollout thus keeps the benefits of self-rollout without its serial cost: apart from a short sink pre-roll (Section~\ref{sec:clean_sink}), the sequential depth drops from $N\times d$ block-level to $d$ stage-level forward passes, and since the supervised pass itself yields the differentiable history, no separate rollout is needed to build it.
Training is therefore faster and more memory-efficient (Table~\ref{tab:training_efficiency}), with a cost nearly independent of block granularity since the sequential depth no longer grows with $N$.

\subsection{Clean Sink as a Persistent Anchor}
\label{sec:clean_sink}

Noisy history suits the local window, which is refreshed at every block and only needs to convey how the scene evolves.
The sink $\Omega$, in contrast, is the only memory that persists beyond the window; it alone anchors subject identity and scene layout and thus calls for a high signal-to-noise ratio (SNR).
We therefore give every stage a clean sink.

Since the rollout stops at the exit stage $d$ (Section~\ref{sec:parallel_rollout}), the cleanest available sink is its own estimate $\widehat{X}_{\Omega}^{(d)}$.
Because every stage needs this estimate, we first roll out the sink blocks sequentially to stage $d$ without gradients, recording their inputs $X_{\Omega}^{(s)}$ for $s\le d$ (Figure~\ref{fig:method_overview}a).
Each stage then prepends the clean sink at timestep zero to its input:
\begin{equation}
    Z^{(s)}
    =
    \bigl[\,
        \underbrace{\operatorname{sg}(\widehat{X}_{\Omega}^{(d)})}_{\text{clean sink},\ t=0}
        \,;\,
        \underbrace{\operatorname{sg}(X^{(s)})}_{\text{all blocks},\ t=t_s}
    \,\bigr],
    \label{eq:sink_input}
\end{equation}
where the sink blocks inside $X^{(s)}$ reuse the recorded $X_{\Omega}^{(s)}$, so each sink block appears both clean and at noise level $t_s$.
Eq.~\ref{eq:aligned_rollout} then takes $Z^{(s)}$ in place of $\operatorname{sg}(X^{(s)})$, and the same forward pass re-encodes the clean sink into $\mathrm{KV}_{\Omega}$ instead of reading a detached cache.
We call this \emph{sink recaching}.
All stages thus share the same clean sink, whose K/V stays in the computation graph at the exit stage.
The history of Section~\ref{sec:aligned_history} thus becomes $\mathcal{H}_i^{(s)}=\mathrm{KV}_{\Omega,<i}\cup\{\mathrm{KV}_j^{(s)}\}_{j\in\mathcal{W}(i)}$, where $\mathrm{KV}_{\Omega,<i}$ is the clean K/V of the sink blocks preceding block $i$ and $\mathcal{W}(i)$ now covers only non-sink predecessors.
For a sink block, $\mathcal{W}(i)$ is empty, so the first sink block has no history.

The sink therefore plays two roles.
As content, its noisy copy is predicted like any other block and receives the DMD loss; as memory, its clean copy supplies $\mathrm{KV}_{\Omega}$ to later blocks and receives the write term of Eq.~\ref{eq:history_grad}.
At inference, only the sink is recached once per video from its final estimate $\widehat{X}_{\Omega}^{(S)}$, while clean-history recaching~\citep{huang2026self,zhuang2026self} re-encodes every block at timestep zero.
This clean anchor stabilizes appearance without suppressing motion, whereas a noisy sink collapses generation to nearly static videos (Section~\ref{sec:ablations}).

\subsection{Streaming Inference}
\label{sec:streaming_inference}

\textbf{Multi-bank streaming.}~To keep latency low, SAF generates the video in a streaming manner at inference, emitting each block once it has passed all $S$ stages instead of running Eq.~\ref{eq:aligned_rollout} over the whole sequence.
Nevertheless, each block should read the history seen in training, namely one at its current noise level.
Multi-bank SAF (SAF$^{\mathrm{M}}$) therefore maintains one history K/V bank $\mathcal{H}^{(s)}$ per stage (Figure~\ref{fig:method_overview}c), which holds exactly $\mathcal{H}_i^{(s)}$ when block $i$ reaches stage $s$:
\begin{equation}
    \bigl(\widehat{X}_i^{(s)}, \mathrm{KV}_{i}^{(s)}\bigr)
    =
    f_\theta\bigl(X_i^{(s)}, t_s; \mathcal{H}^{(s)}, p_i\bigr),
    \qquad
    \mathcal{H}^{(s)} \leftarrow \operatorname{Commit}_L\bigl(\mathcal{H}^{(s)}, \mathrm{KV}_{i}^{(s)}\bigr),
    \label{eq:multibank_inference}
\end{equation}
where $p_i$ is the prompt of block $i$, and $\operatorname{Commit}_L$ appends the new K/V after the forward pass and evicts the oldest non-sink entries beyond the window.
All banks share the same clean sink.
Because the K/V computed during denoising is reused as history, each block needs only $S$ forward passes rather than the $S{+}1$ required by clean-history recaching~\citep{huang2026self,zhuang2026self}.

\textbf{Multi-GPU stage pipelining.}~Since block $i$ at stage $s$ depends only on its latent from stage $s{-}1$ and $\mathcal{H}^{(s)}$, stages decouple across GPUs: GPU $s$ holds a copy of $f_\theta$ and $\mathcal{H}^{(s)}$, denoises blocks in order at stage $s$, and passes each latent to GPU $s{+}1$ while keeping the K/V locally (Figure~\ref{fig:method_overview}c and Algorithm~\ref{alg:pipeline_saf}).
While GPU $s$ processes block $i$, GPU $s{+}1$ processes block $i{-}1$.
Once the pipeline is full, $S$ blocks are in flight, each at a different stage, and each GPU stores one history K/V bank.
Clean-history methods~\citep{huang2026self,zhuang2026self} cannot be pipelined this way, since block $i{+}1$ must wait until block $i$ completes all stages and is recached at $t=0$.

\textbf{Single-bank streaming.}~When memory is limited, single-bank SAF (SAF$^{\mathrm{S}}$) runs the same model on one GPU with a single bank $\mathcal{H}$ shared by all stages (Figure~\ref{fig:method_overview}c and Algorithm~\ref{alg:single_bank}).
Once a block is fully denoised, only the K/V from one of its stages is committed to $\mathcal{H}$, with that stage's noise level either fixed to $t$ (Fixed$t$) or sampled per block.
Since noisier history K/V favors dynamics and cleaner history K/V favors fidelity (Figure~\ref{fig:motivation_stage}), SAF$^{\mathrm{S}}$ balances the two by sampling noise levels $(250,500,750)$ with probabilities $(0.5,0.25,0.25)$, which we call Mix.
This variant saves memory but cannot be pipelined, and most stages read history at a noise level mismatched with training, so SAF$^{\mathrm{M}}$ remains our default.

\newsavebox{\algboxA}\newsavebox{\algboxB}\newlength{\algboxH}
\newcommand{\algpanel}[1]{\begin{minipage}[t]{0.485\textwidth}\vspace{0pt}\noindent\usebox{#1}\par
\vspace{\dimexpr\algboxH-\ht#1-\dp#1\relax}\vspace{4pt}\hrule height 0.8pt\end{minipage}}
\begin{figure*}[t]
\centering

\sbox{\algboxA}{\begin{minipage}[t]{0.485\textwidth}
\vspace{0pt}

\hrule height 0.8pt
\vspace{2pt}
\captionsetup{type=algorithm,font=small}
\caption{Parallel Training of SAF}
\label{alg:aligned_saf}
\vspace{-4pt}
\hrule height 0.4pt
\vspace{4pt}

\fontsize{7.6pt}{8.8pt}\selectfont
\begin{algorithmic}[1]
\Require Denoiser $f_\theta$, stages $t_{1:S}$, sampler $\Psi$, $N$ blocks,
causal mask $M_{\mathrm{causal}}$, sink $\Omega$, prompt $p$
\Ensure One training iteration

\State Sample exit stage
$d\sim\mathcal{U}\{1,\ldots,S\}$
\State Sample $X^{(1)},\epsilon_{1:d-1}\sim\mathcal N(0,I)$
\For{sink block $i\in\Omega$ in order} \Comment{no gradients}
    \For{$s=1,\ldots,d$}
        \State $\widehat X_i^{(s)}\gets f_\theta(X_i^{(s)},t_s;\mathcal H_i^{(s)},p)$
        \State $X_i^{(s+1)}\gets\Psi_s(\widehat X_i^{(s)},\epsilon_s)$ if $s<d$
    \EndFor
\EndFor

\For{$s=1,\ldots,d$}
    \State $Z^{(s)}\gets\bigl[\operatorname{sg}(\widehat X_{\Omega}^{(d)});\,\operatorname{sg}(X^{(s)})\bigr]$
    \Statex \hspace{\algorithmicindent}
    \emph{clean sink at $t=0$, all blocks at $t_s$}
    \If{$s<d$} \Comment{no gradients}
        \State
        $\widehat X^{(s)}
        \gets
        f_\theta(
        Z^{(s)},
        t_s;
        M_{\mathrm{causal}},
        p)$
        \State
        $X^{(s+1)}
        \gets
        \Psi_s(
        \widehat X^{(s)},
        \epsilon_s)$
        \Statex \hspace{\algorithmicindent}\hspace{\algorithmicindent}
        with sink entries reset to $X_{\Omega}^{(s+1)}$
    \Else \Comment{with gradients}
        \State
        $\widehat X^{(d)}
        \gets
        f_\theta(
        Z^{(d)},
        t_d;
        M_{\mathrm{causal}},
        p)$
        \Statex \hspace{\algorithmicindent}\hspace{\algorithmicindent}
        \emph{sink and history K/V remain differentiable}
    \EndIf
\EndFor

\State
$\mathcal L
\gets
\mathcal L_{\mathrm{DMD}}
(\widehat X^{(d)},p)$
\State Backpropagate the read and write terms; update $\theta$
\State Update the score models as in DMD

\end{algorithmic}
\end{minipage}}%
%
\sbox{\algboxB}{\begin{minipage}[t]{0.485\textwidth}
\vspace{0pt}

\hrule height 0.8pt
\vspace{2pt}
\captionsetup{type=algorithm,font=small}
\caption{Multi-GPU Pipeline Inference of SAF$^{\mathrm{M}}$}
\label{alg:pipeline_saf}
\vspace{-4pt}
\hrule height 0.4pt
\vspace{4pt}

\fontsize{7.6pt}{8.8pt}\selectfont
\begin{algorithmic}[1]
\Require Denoiser $f_\theta$, stages $t_{1:S}$, sampler $\Psi$, $N$ blocks, window $L$, sink $\Omega$, block prompts $p_{1:N}$, $S$ GPUs
\Ensure Ordered clean blocks $\widehat X_{1:N}^{(S)}$

\State Disable gradients and replicate $f_\theta$ across GPUs
\State Generate and emit the sink blocks sequentially;
\Statex encode $\widehat X_{\Omega}^{(S)}$ once at $t=0$ to obtain $\mathrm{KV}_{\Omega}$
\State Each GPU $s$ initializes
$\mathcal H^{(s)}\gets\operatorname{Init}(\mathrm{KV}_{\Omega})$

\For{clock $r=1,\ldots,N-|\Omega|+S-1$}
    \ForAll{GPUs $s$ in parallel}
        \State $i\gets|\Omega|+r-s+1$; skip if $i\notin(|\Omega|,N]$

        \If{$s=1$}
            \State Sample
            $X_i^{(1)}\sim\mathcal N(0,I)$
        \Else
            \State Receive
            $X_i^{(s)}$
            from GPU $s-1$
        \EndIf

        \State
        $(\widehat X_i^{(s)},
        \mathrm{KV}_{i}^{(s)})
        \gets
        f_\theta(
        X_i^{(s)},
        t_s;
        \mathcal H^{(s)},
        p_i)$

        \State
        $\mathcal H^{(s)}
        \gets
        \operatorname{Commit}_L(
        \mathcal H^{(s)},
        \mathrm{KV}_{i}^{(s)})$

        \If{$s<S$}
            \State Sample
            $\epsilon_{i,s}\sim\mathcal N(0,I)$
            \State
            $X_i^{(s+1)}
            \gets
            \Psi_s(
            \widehat X_i^{(s)},
            \epsilon_{i,s})$
            \State Send
            $X_i^{(s+1)}$
            to GPU $s+1$
        \Else
            \State Emit
            $\widehat X_i^{(S)}$
        \EndIf

    \EndFor
\EndFor

\end{algorithmic}
\end{minipage}}%
\setlength{\algboxH}{\dimexpr\ht\algboxA+\dp\algboxA\relax}%
\ifdim\dimexpr\ht\algboxB+\dp\algboxB\relax>\algboxH\setlength{\algboxH}{\dimexpr\ht\algboxB+\dp\algboxB\relax}\fi
\algpanel{\algboxA}\hfill\algpanel{\algboxB}

\end{figure*}

\section{Experiments}
\label{sec:experiments}

\subsection{Experimental Setup}

Our models are trained on 5-second, single-prompt video clips in both framewise and chunkwise settings, and we evaluate whether this short-horizon training transfers to longer and interactive generation. 
Specifically, three complementary benchmarks are used: \textbf{VBench}~\citep{huang2024vbench} evaluates 5-second clips at the training horizon, \textbf{Interactive}~\citep{yang2025longlive,ji2025memflow} generates 60-second videos across six 10-second segments, each conditioned on a distinct prompt, and \textbf{MovieGen-100s}~\citep{polyak2025moviegen,zhang2026unitemp} generates 100-second videos from 128 prompts to test generation quality at $20{\times}$ the training length.
We compare with Self-Forcing~\citep{huang2026self}, LongLive~\citep{yang2025longlive}, Causal-Forcing~\citep{zhu2026causal}, HiAR~\citep{zou2026hiar}, and SGF~\citep{zhuang2026self}, re-evaluating all of them with official weights under identical random seeds; only Causal-Forcing and SGF release framewise checkpoints.  Additional protocol and scoring details are provided in Appendix~\ref{sec:evaluation_details}.
\begin{table}[tbp]
\centering
\caption{\textbf{Chunkwise evaluation with existing methods.} FPS is measured on H100 GPUs; when two throughput values are reported, the second denotes 4-GPU pipelined parallel inference. Best results are \textbf{bold} and second-best results are \underline{underlined}.}
\label{tab:main_evaluation}
\setlength{\tabcolsep}{2pt}
\renewcommand{\arraystretch}{1.03}
\begin{adjustbox}{width=0.9\linewidth,center}
\begin{tabular}{lccccccc}
        \toprule
        \multirow{2}{*}{Method}
        & \multirow{2}{*}{\shortstack{Throughput\\(FPS) $\uparrow$}}
        & \multicolumn{3}{c}{VBench}
        & \multicolumn{2}{c}{Interactive}
        & \multicolumn{1}{c}{MovieGen 100s} \\
        \cmidrule(lr){3-5}
        \cmidrule(lr){6-7}
        \cmidrule(lr){8-8}
        & & Total $\uparrow$ & Quality $\uparrow$ & Semantic $\uparrow$ & ViCLIP $\uparrow$ & Quality $\uparrow$ & Quality $\uparrow$ \\
        \midrule
        Self-Forcing & $17.0$ & $83.80$ & $84.59$ & $80.64$ & $21.88$ & $80.69$ & $78.74$ \\
        LongLive & $20.7$ & $83.22$ & $83.68$ & $\mathbf{81.37}$ & $\mathbf{25.74}$ & $82.17$ & $82.83$ \\
        Causal-Forcing & $17.0$ & $\underline{84.88}$ & $\underline{85.93}$ & $80.67$ & $24.21$ & $81.66$ & $79.97$ \\
        HiAR & $11.2$ / $\underline{30.0}$ & $82.66$ & $83.34$ & $79.93$ & $24.16$ & $80.97$ & $81.84$ \\
        SGF & $20.7$ & $84.76$ & $85.74$ & $\underline{80.83}$ & $24.63$ & $83.83$ & $\underline{84.63}$ \\
        \cmidrule(lr){1-8}
        \textbf{SAF}$^{\mathrm{S}}$ & $\mathbf{22.9}$ & $84.61$ & $85.80$ & $79.87$ & $24.70$ & $\underline{84.29}$ & $84.34$ \\
        \textbf{SAF}$^{\mathrm{M}}$ & $\underline{22.8}$ / $\mathbf{49.1}$ & $\mathbf{85.04}$ & $\mathbf{86.30}$ & $80.00$ & $\underline{25.34}$ & $\mathbf{85.59}$ & $\mathbf{85.00}$ \\
        \bottomrule
    \end{tabular}
\end{adjustbox}
\end{table}

\subsection{Comparison with Existing Methods}

\textbf{Quantitative analysis.}
Table~\ref{tab:main_evaluation} reports the chunkwise comparison, where ours with SAF$^{\mathrm{M}}$ leads nearly all aggregate metrics across short-video, interactive, and long-video generation.
LongLive attains a higher Interactive ViCLIP score, plausibly reflecting its dedicated prompt-switching and long-video training, but ours still achieves higher Interactive Quality and MovieGen-100s Quality.
The only remaining deficit is a small drop in VBench Semantic, a metric that often decreases when motion and dynamics are stronger.
Ours is also the fastest single-GPU method because it avoids the extra timestep-zero forward pass required by clean-history recaching, and its stage-wise design enables four-GPU pipelined inference for a further throughput gain.
SAF$^{\mathrm{S}}$ is slightly weaker than SAF$^{\mathrm{M}}$ due to its train--test mismatch, yet it remains competitive and preserves clear quality advantages.
Full dimension-wise scores and the framewise comparison are provided in Appendix~\ref{sec:full_results}.

\newsavebox{\ablationtabbox}
\newlength{\ablationtabheight}
\begin{table}[!ht]
\centering
\captionsetup{font=footnotesize,skip=3pt}
\begin{minipage}[t]{0.485\textwidth}
\captionsetup{type=table}
\caption{\textbf{Component and K/V bank ablations on 20-second Interactive generation.} At inference, SAF$^{\mathrm{M}}$ uses stage-aligned multiple K/V banks and SAF$^{\mathrm{S}}$ uses one bank written at noise levels sampled from $(250,500,750)$ with probabilities $(0.5,0.25,0.25)$. Indented w/o rows: training ablations under the same inference; Fixed$t$: SAF with one bank written at noise level $t$. Best per group in \textbf{bold}, second \underline{underlined}.}
\label{tab:inference_ablation}
\end{minipage}\hfill
\begin{minipage}[t]{0.485\textwidth}
\captionsetup{type=figure}
\caption{\textbf{Quality across SDF training schedules on the first 20 seconds of MovieGen-100s.} The x-axis gives the random/fixed/aligned history ratio in training ($0/0/100$ is SAF; Appendix~\ref{sec:sdf}). Bars are inference policies: Aligned uses stage-aligned multiple K/V banks (SAF$^{\mathrm{M}}$), Mix one bank written at noise levels sampled from $(250,500,750)$ with probabilities $(0.5,0.25,0.25)$ (SAF$^{\mathrm{S}}$), and Fixed250 one bank written at noise level 250.}
\label{fig:mode_ratio_quality}
\end{minipage}\par\vspace{3pt}
\sbox{\ablationtabbox}{%
\fontsize{6.5}{7.6}\selectfont
\setlength{\tabcolsep}{2.4pt}%
\renewcommand{\arraystretch}{1.0}%
\begin{adjustbox}{max width=0.485\textwidth}
\begin{tabular}{lccccc}
\toprule
Method & ViCLIP $\uparrow$ & Quality $\uparrow$ & Smooth $\uparrow$ & Dyn. $\uparrow$ & Img. $\uparrow$ \\
\midrule
\textbf{SAF}$^{\mathrm{M}}$ & \textbf{26.54} & \textbf{85.18} & 98.93 & \textbf{64.10} & \underline{71.82} \\
\quad w/o K/V grad. & \underline{26.15} & \underline{84.64} & \underline{99.07} & \underline{51.30} & \textbf{72.58} \\
\quad w/o sink recache & 24.42 & 81.70 & \textbf{99.48} & 6.90 & 71.24 \\
\midrule
\textbf{SAF}$^{\mathrm{S}}$ & \underline{25.99} & \textbf{84.22} & 99.06 & \textbf{48.00} & \underline{72.23} \\
\quad w/o K/V grad. & \textbf{26.01} & \underline{84.02} & \underline{99.07} & \underline{44.50} & \textbf{72.72} \\
\quad w/o sink recache & 24.21 & 81.30 & \textbf{99.41} & 4.30 & 71.04 \\
\midrule
Fixed750 & \textbf{25.71} & \textbf{84.68} & 98.94 & \textbf{55.60} & 71.93 \\
Fixed500 & 25.40 & \underline{84.12} & \underline{99.02} & \underline{47.10} & \underline{72.13} \\
Fixed250 & \underline{25.65} & 83.99 & \textbf{99.10} & 45.70 & \textbf{72.36} \\
\bottomrule
\end{tabular}
\end{adjustbox}}%
\setlength{\ablationtabheight}{\dimexpr\ht\ablationtabbox+\dp\ablationtabbox\relax}%
\begin{minipage}[t]{0.485\textwidth}
\vspace{0pt}
\centering
\usebox{\ablationtabbox}
\end{minipage}\hfill
\begin{minipage}[t]{0.485\textwidth}
\vspace{0pt}
\centering
\includegraphics[width=\linewidth,height=\ablationtabheight,keepaspectratio]{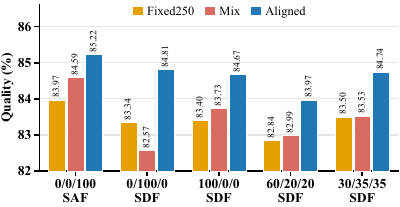}
\end{minipage}
\end{table}

\begin{figure}[!ht]
\centering
\includegraphics[width=0.9\linewidth]{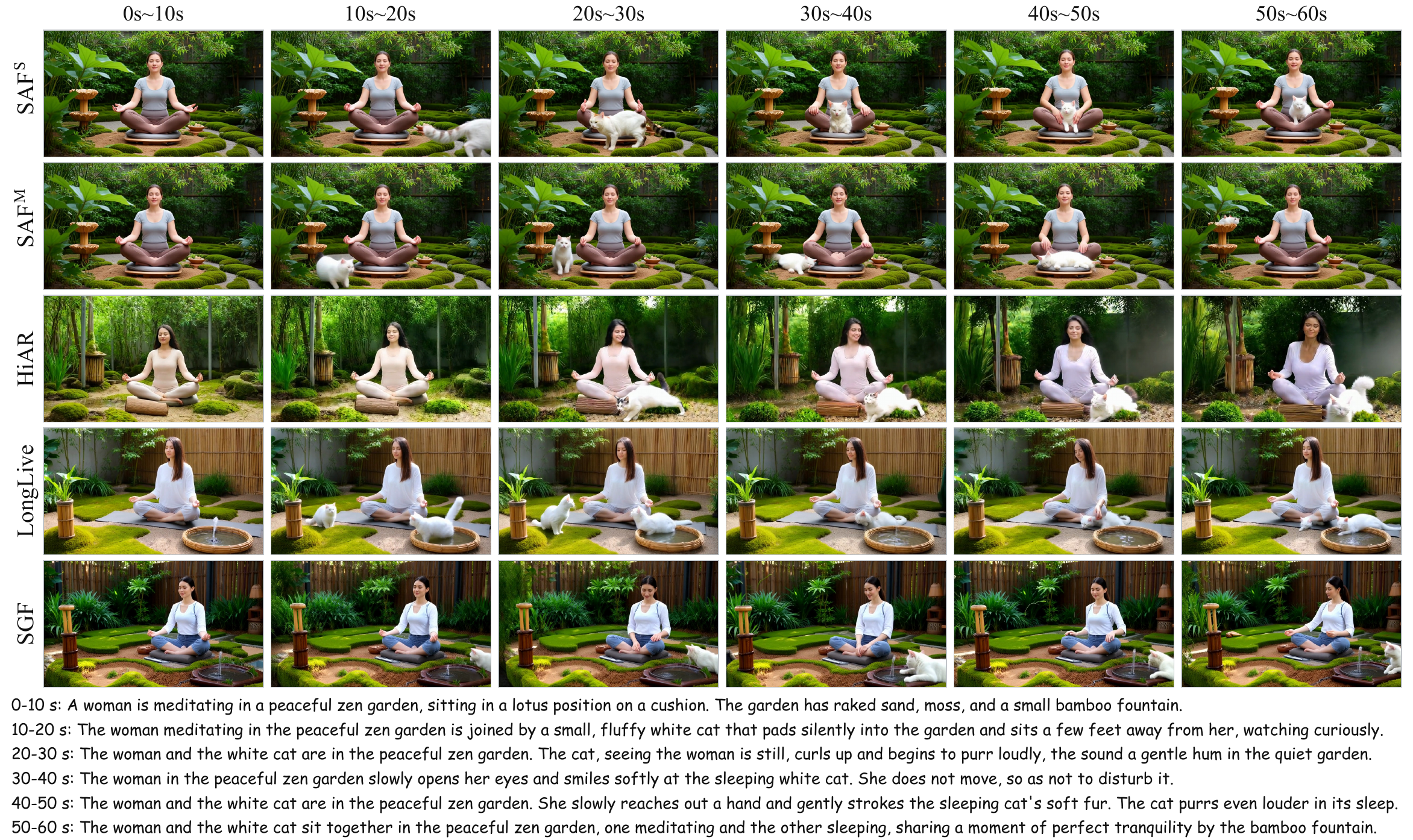}
\captionsetup{skip=3pt}
\caption{\textbf{Chunkwise Interactive qualitative comparison.} Ours tracks the changing ten-second prompts with faithful semantics and stable visual quality, while competing methods exhibit quality degradation, object-count errors, or missed interactions.}
\label{fig:qual_interactive}
\end{figure}

\begin{figure}[!ht]
\centering
\includegraphics[width=0.9\linewidth]{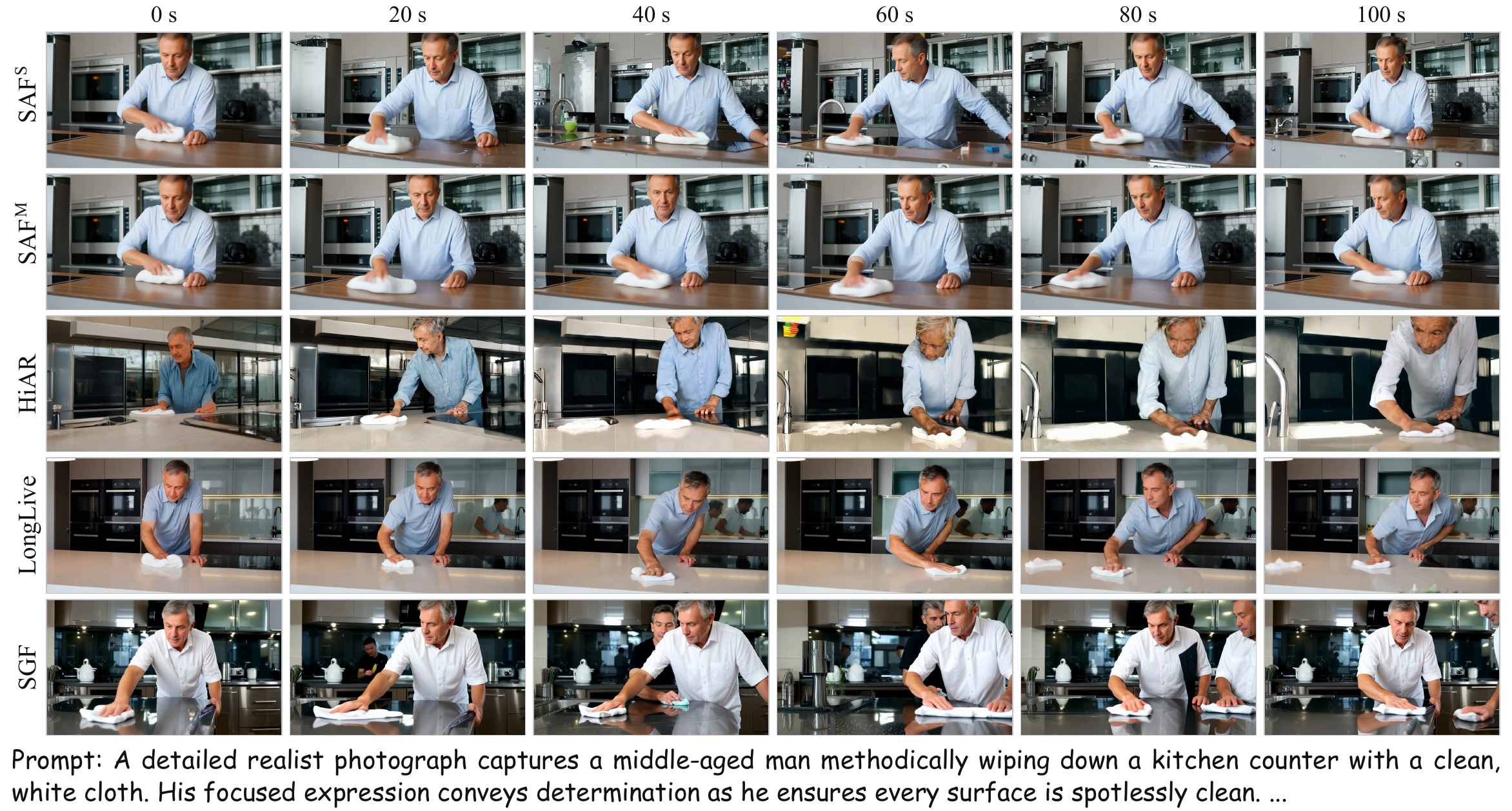}
\captionsetup{skip=3pt}
\caption{\textbf{Chunkwise MovieGen-100s qualitative comparison.} Ours sustains the long-horizon table-wiping action with stable scene structure, whereas competing methods accumulate artifacts, duplicate subjects, or produce inconsistent reflections.}
\label{fig:qual_movie100}
\end{figure}

\textbf{Qualitative analysis.}
Figures~\ref{fig:qual_interactive} and~\ref{fig:qual_movie100} show that the chunkwise gains translate into stronger semantic alignment and visual stability.
In Interactive generation, ours tracks successive prompt changes while preserving visual fidelity; in contrast, LongLive can introduce semantic errors such as duplicating the cat, and SGF fails to realize the requested prompt-conditioned motion.
On MovieGen-100s, ours maintains the table-wiping motion and scene layout over the full horizon, whereas SGF duplicates the human subject and LongLive produces inconsistent background reflections.
Additional comparisons, including framewise results, are provided in Appendix~\ref{sec:qualitative_gallery}, and more visualizations are provided in the supplementary material.


\subsection{Ablation Studies}
\label{sec:ablations}

We conduct ablations on 20-second subsets of MovieGen-100s and Interactive, with protocol details and full metrics reported in Appendix~\ref{sec:ablation_details}.
Table~\ref{tab:inference_ablation} shows consistent trends for both SAF$^{\mathrm{M}}$ and SAF$^{\mathrm{S}}$.
Detaching the history K/V mildly reduces Quality but substantially lowers Dynamic Degree, confirming the importance of the write term, i.e., future-to-history supervision.
Removing sink recaching causes the largest degradation in both metrics.
For single-bank inference under Fixed$t$, a larger $t$ generally improves Quality and Dynamic Degree, but reduces Motion Smoothness and Imaging Quality, revealing a trade-off between dynamics and visual stability.
We therefore adopt the Mix policy for SAF$^{\mathrm{S}}$, which provides a more balanced operating point across dynamics, smoothness, and image fidelity.

To examine how the noise level of the history K/V affects training, Figure~\ref{fig:mode_ratio_quality} compares schedules of Self-Diffusion Forcing (SDF), a two-pass generalization of SAF that writes each block's K/V at a random, fixed ($t{=}250$), or aligned stage (Appendix~\ref{sec:sdf}).
A schedule $a/b/c$ samples these three policies with probabilities $a\%$, $b\%$, and $c\%$, so $0/0/100$ is SAF.
Across the evaluated schedules, aligned inference consistently achieves the highest Quality, while Mix is typically second-best.
For the fixed-trained model, Fixed250 performs better, consistent with its matched training history distribution.
SAF ($0/0/100$) achieves the highest Quality under every inference policy.

\par\vspace{4pt}\noindent
{\captionsetup{font=footnotesize,skip=3pt}%
\begin{minipage}[b]{0.54\textwidth}
\textbf{Training efficiency.}
SAF is the most efficient in both settings (Table~\ref{tab:training_efficiency}).
Block-parallel rollout reduces the sequential depth to the number of stages, and the supervised forward pass itself yields the differentiable history, removing the history-construction rollout and per-block recaching.
The gain is largest in the framewise setting, where serial rollout is most costly: SAF trains $1.8\times$ faster than SGF and nearly matches its own chunkwise speed (137\,s vs.\ 133\,s).
Even the two-pass SDF (Appendix~\ref{sec:sdf}) is slightly faster than SGF, as it skips per-block timestep-zero recaching.
\par\vspace{0pt}
\end{minipage}\hfill
\begin{minipage}[b]{0.435\textwidth}
\centering
\captionsetup{type=table}
\caption{\textbf{Training efficiency on one GB300} at batch size 4: memory (GiB) and time (s) per five-step cycle. Local attention spans 12 frames (chunkwise) and 21 frames (framewise). Best is \textbf{bold} and second-best \underline{underlined}.}
\label{tab:training_efficiency}
\scriptsize
\setlength{\tabcolsep}{3pt}
\renewcommand{\arraystretch}{1.0}
\begin{adjustbox}{max width=\linewidth}
\begin{tabular}{lcccc}
\toprule
\multirow{2}{*}{Method} & \multicolumn{2}{c}{Chunkwise} & \multicolumn{2}{c}{Framewise} \\
\cmidrule(lr){2-3}\cmidrule(lr){4-5}
& Mem. $\downarrow$ & Time $\downarrow$ & Mem. $\downarrow$ & Time $\downarrow$ \\
\midrule
        LongLive & 173.63 & \underline{158.32} & 174.21 & \underline{232.64} \\
        SGF & 187.45 & 186.73 & 187.45 & 249.21 \\
\midrule
        SDF & \underline{160.85} & 184.33 & \underline{171.30} & 245.87 \\
        \textbf{SAF} & \textbf{155.18} & \textbf{132.78} & \textbf{164.79} & \textbf{137.04} \\
\bottomrule
\end{tabular}
\end{adjustbox}
\par\vspace{0pt}
\end{minipage}}\par


\section{Conclusion}
\label{sec:conclusion}

In this paper, we show that history K/V along the denoising trajectory, when aligned with the current noise level, can serve as an effective and trainable memory for streaming video diffusion.
\methodname realizes this by aligning each block's history with its denoising stage and rolling out all blocks at each stage in a single causal forward pass.
This keeps the noisy history differentiable, allowing future losses to supervise how it is written, while avoiding both a separate no-gradient rollout to construct the history and per-block timestep-zero recaching.
Empirically, SAF improves long-horizon generation and prompt switching, accelerates training by up to $1.8\times$ over SGF with lower memory, and supports efficient multi-GPU pipelined inference.
Its noisy local history can still cause abrupt changes when new content enters the frame under camera motion (Appendix~\ref{sec:limitations}).
Overall, our results suggest that optimizing history directly along the denoising trajectory is a promising path toward efficient, long-horizon, and interactive video generation.

\clearpage
\bibliography{main}
\bibliographystyle{iclr2027_conference}

\clearpage
\appendix
\raggedbottom
\renewcommand{\floatpagefraction}{0.85}
\renewcommand{\bottomfraction}{0.7}

\section{Self-Diffusion Forcing}
\label{sec:sdf}

\begin{figure}[t]
\centering
\includegraphics[width=\textwidth]{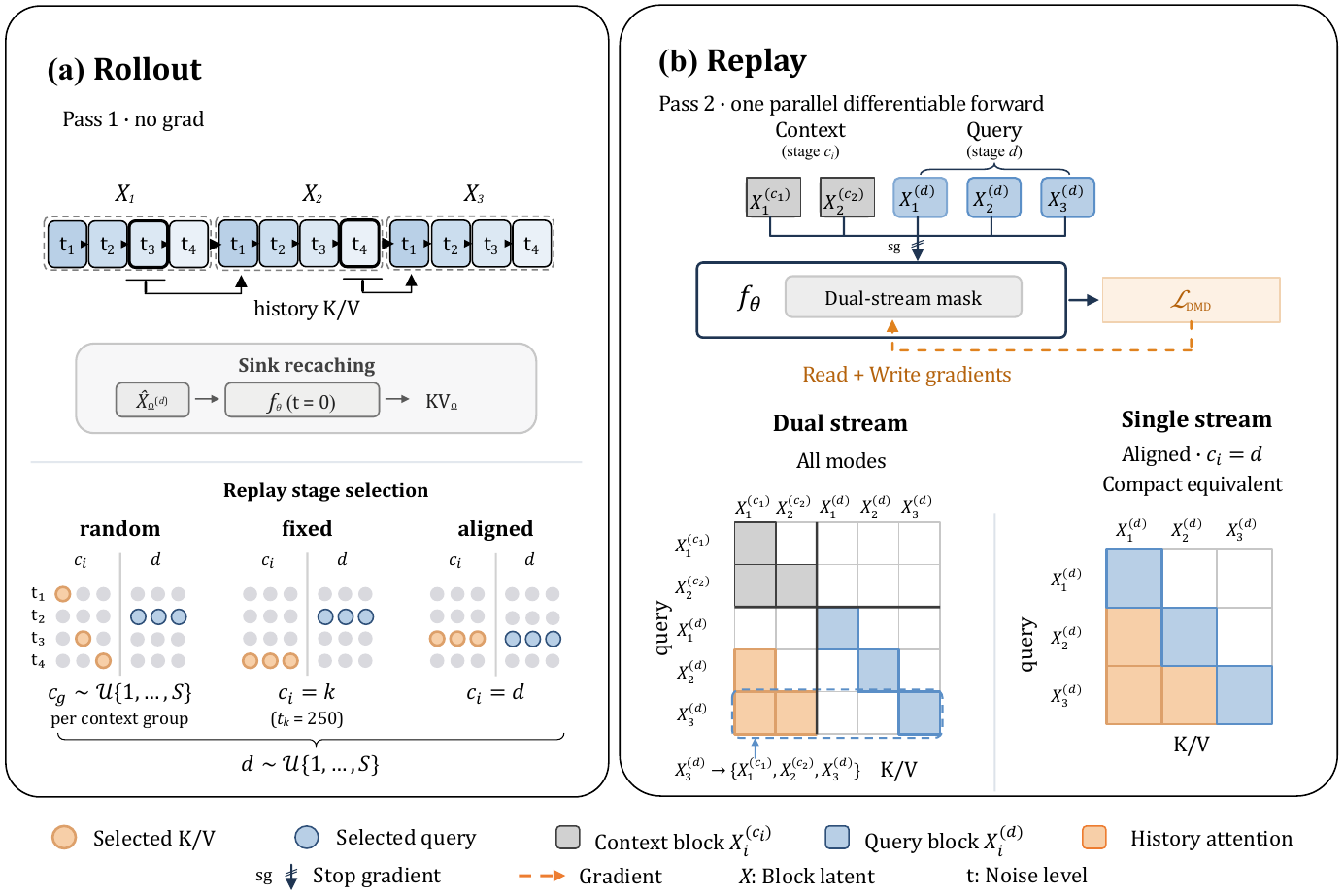}
\captionsetup{skip=4pt}
\caption{\textbf{Two-pass training of SDF.}
\textbf{(a)} Pass~1 rolls out all blocks serially without gradients, recaches the sink from $\widehat X_{\Omega}^{(d)}$, and records the trajectory; the history of each block is read at the stage chosen by the random, fixed, or aligned policy.
\textbf{(b)} Pass~2 replays the recorded states in one differentiable forward pass over a context stream $C$ and a query stream $D$, so the DMD loss on the queries reaches the context K/V they attend to.
Under the aligned policy ($c_i=d$), context and query coincide and the two streams collapse into one.
For clarity, the attention masks omit the clean sink; as in SAF, block $i$ also attends to the clean K/V of the preceding sink blocks, $\mathrm{KV}_{\Omega,<i}$ (Section~\ref{sec:clean_sink}).
At the top of (a), $X_i$ is block $i$, whose K/V enters the history at the bold stage; the context and query streams in (b) hold its recorded states $X_i^{(c_i)}$ and $X_i^{(d)}$, and $c_g$ is shared by context group $g$ (Eq.~\ref{eq:history_policies}).}
\label{fig:sdf_framework}
\end{figure}

SAF encodes the history of every block at the stage of the current block.
To test this choice, our schedule ablations (Section~\ref{sec:ablations} and Appendix~\ref{sec:schedule_ablations}) use Self-Diffusion Forcing (SDF), a generalization that decouples the stage $c_i$ at which block $i$ writes its K/V into the history from the query stage $d\sim\mathcal{U}\{1,\ldots,S\}$ at which predictions are supervised.
In SAF, $d$ is also the exit stage at which the rollout stops (Section~\ref{sec:parallel_rollout}); in SDF, it does not truncate the rollout: Pass~1 runs all $S$ stages, and $d$ only selects which recorded states are replayed with gradients.
Figure~\ref{fig:sdf_framework} illustrates the procedure.

\paragraph{History-stage policies.}
SDF supports three policies for the history stage:
\begin{equation}
    c_i = d\ \text{(aligned)},
    \qquad
    c_i = k\ \text{(fixed)},
    \qquad
    c_i = c_{g(i)},\ c_g\sim\mathcal{U}\{1,\ldots,S\}\ \text{(random)},
    \label{eq:history_policies}
\end{equation}
where $k$ is the stage with $t_k=250$ and $g(i)$ is the group of three latent frames containing block $i$.
A training schedule draws one policy per sample, and schedule $a/b/c$ uses the random, fixed, and aligned policies with probabilities $a\%$, $b\%$, and $c\%$; the $0/0/100$ schedule is SAF.

\paragraph{Two-pass training.}
When $c_i\neq d$, the history and the query of a block lie at different noise levels and cannot share a forward pass, so SDF trains in two passes.
As in SAF, the sink blocks are first rolled out sequentially, and their estimate $\widehat X_{\Omega}^{(d)}$ at the query stage $d$ serves as the clean sink recached in both passes.
Pass~1 rolls out the remaining blocks serially without gradients, reading the history at the policy stage, and records their states.
Pass~2 packs the clean sink, a context stream $C$ with $C_i=X_i^{(c_i)}$, and a query stream $D$ with $D_i=X_i^{(d)}$, all recorded in Pass~1, into one differentiable forward pass.
Under the dual-stream mask $M_{\mathrm{dual}}$, query block $i$ attends to the clean sink blocks before it, the context blocks $j<i$ within the window, and its own query tokens, but never to its own context copy, other query blocks, or future context.
Only query predictions receive the DMD loss, and its gradients reach the sink and context K/V that the queries attend to, but not the recorded trajectory.
Under the aligned policy, $C$ and $D$ coincide and $M_{\mathrm{dual}}$ reduces to $M_{\mathrm{causal}}$.
SAF goes one step further: because every stage produces its own history, it replaces the serial Pass~1 with the block-parallel rollout of Section~\ref{sec:parallel_rollout}.
Algorithm~\ref{alg:general_sdf} in Appendix~\ref{sec:algorithm_details} gives the full procedure.

\section{Algorithm Details}
\label{sec:algorithm_details}

This section complements Algorithms~\ref{alg:aligned_saf} and~\ref{alg:pipeline_saf} with the shared cache operations, single-GPU inference for SAF$^{\mathrm M}$ and SAF$^{\mathrm S}$ (Algorithms~\ref{alg:serial_multibank} and~\ref{alg:single_bank}), and the two-pass SDF training of Appendix~\ref{sec:sdf} (Algorithm~\ref{alg:general_sdf}).
Notation follows Section~\ref{sec:method}.

\paragraph{Denoiser and sampling.}
$\mathrm{KV}_i^{(s)}$ collects the K/V of block $i$ from all attention layers.
In the training algorithms, $f_\theta$ processes all blocks in one forward pass under the indicated attention mask, whereas at inference it processes one block against a history K/V bank.
Gaussian draws are indexed by sample, block, and stage, so all execution schedules of the same sample use identical noise.
Blocks keep their original temporal indices for positional encoding.
Training uses a single prompt $p$, whereas at inference block $i$ uses its own prompt $p_i$, so a prompt switch changes only the text conditioning and leaves the generated history intact.

\paragraph{Cache operations.}
The sink $\Omega$ consists of the leading blocks of the video, and \emph{sink recaching} encodes their clean latents at $t=0$ into the clean K/V $\mathrm{KV}_{\Omega}$.
During training, the sink is rolled out to the exit stage $d$, and the parallel forward pass of every stage recaches it from the same estimate $\widehat X_{\Omega}^{(d)}$ (Eq.~\ref{eq:sink_input}); SDF likewise uses $\widehat X_{\Omega}^{(d)}$ (Appendix~\ref{sec:sdf}).
At inference, the sink blocks are generated first and recached once from their final estimate $\widehat X_{\Omega}^{(S)}$, after which the remaining blocks are streamed.
$\operatorname{Init}(\mathrm{KV}_{\Omega})$ creates a bank that holds $\mathrm{KV}_{\Omega}$ and an empty local window, and all stage banks share the same sink.
$\operatorname{Commit}_L(\mathcal H,\mathrm{KV})$ appends the K/V of the current block to $\mathcal H$ after its forward pass and evicts the oldest non-sink entries beyond the window $L$; the sink is never evicted.
Apart from sink recaching, no procedure below runs a forward pass at $t=0$.

\paragraph{Single-GPU multi-bank inference.}
Algorithm~\ref{alg:serial_multibank} runs SAF$^{\mathrm M}$ on one GPU by denoising each block through all $S$ stages before starting the next.
It respects the same dependencies as the pipeline of Algorithm~\ref{alg:pipeline_saf}, so both produce identical videos from the same noise: when block $i$ reaches stage $s$, bank $\mathcal H^{(s)}$ holds exactly its history $\mathcal H_i^{(s)}$.

\begin{algorithm}[!t]
\caption{Single-GPU Multi-Bank Inference of SAF$^{\mathrm M}$}
\label{alg:serial_multibank}
\footnotesize
\begin{algorithmic}[1]
\Require Denoiser $f_\theta$, stages $t_{1:S}$, sampler $\Psi$, $N$ blocks, window $L$, sink $\Omega$, block prompts $p_{1:N}$
\Ensure Ordered clean blocks $\widehat X_{1:N}^{(S)}$
\State Disable gradients; generate and emit the sink blocks sequentially
\State Encode $\widehat X_{\Omega}^{(S)}$ once at $t=0$ to obtain $\mathrm{KV}_{\Omega}$
\State $\mathcal H^{(s)}\gets\operatorname{Init}(\mathrm{KV}_{\Omega})$ for $s=1,\ldots,S$
\For{$i=|\Omega|+1,\ldots,N$}
 \State Sample $X_i^{(1)}\sim\mathcal N(0,I)$
 \For{$s=1,\ldots,S$}
  \State $(\widehat X_i^{(s)},\mathrm{KV}_i^{(s)})\gets f_\theta(X_i^{(s)},t_s;\mathcal H^{(s)},p_i)$
  \State $\mathcal H^{(s)}\gets\operatorname{Commit}_L(\mathcal H^{(s)},\mathrm{KV}_i^{(s)})$
  \If{$s<S$}
   \State Sample $\epsilon_{i,s}\sim\mathcal N(0,I)$; $X_i^{(s+1)}\gets\Psi_s(\widehat X_i^{(s)},\epsilon_{i,s})$
  \Else
   \State Emit $\widehat X_i^{(S)}$
  \EndIf
 \EndFor
\EndFor
\end{algorithmic}
\end{algorithm}

\paragraph{Single-bank inference.}
Algorithm~\ref{alg:single_bank} keeps one bank $\mathcal H$ shared by all stages.
For each block, it samples the stage $c_i\sim\pi$ whose K/V is committed, where $\pi$ places all mass on the stage with noise level $t$ for Fixed$t$, and for Mix samples the stages with noise levels $(250,500,750)$ with probabilities $(0.5,0.25,0.25)$.
All stages of block $i$ read the same history, and only the K/V from stage $c_i$ is committed once the block is fully denoised.
The bank thus holds one entry per block, and different blocks may be written at different noise levels.

\begin{algorithm}[!t]
\caption{Single-Bank Inference of SAF$^{\mathrm S}$}
\label{alg:single_bank}
\footnotesize
\begin{algorithmic}[1]
\Require Denoiser $f_\theta$, stages $t_{1:S}$, sampler $\Psi$, $N$ blocks, window $L$, sink $\Omega$, block prompts $p_{1:N}$
\Require Commit policy $\pi$ over stages $\{1,\ldots,S\}$ (Fixed$t$ or Mix)
\Ensure Ordered clean blocks $\widehat X_{1:N}^{(S)}$
\State Disable gradients; generate and emit the sink blocks sequentially
\State Encode $\widehat X_{\Omega}^{(S)}$ once at $t=0$ to obtain $\mathrm{KV}_{\Omega}$
\State $\mathcal H\gets\operatorname{Init}(\mathrm{KV}_{\Omega})$
\For{$i=|\Omega|+1,\ldots,N$}
 \State Sample $c_i\sim\pi$ and $X_i^{(1)}\sim\mathcal N(0,I)$; $\mathrm{KV}_i^\star\gets\emptyset$
 \For{$s=1,\ldots,S$}
  \State $(\widehat X_i^{(s)},\mathrm{KV}_i^{(s)})\gets f_\theta(X_i^{(s)},t_s;\mathcal H,p_i)$ \Comment{$\mathcal H$ is read-only}
  \If{$s=c_i$}
   \State $\mathrm{KV}_i^\star\gets\mathrm{KV}_i^{(s)}$ \Comment{keep the K/V of stage $c_i$}
  \EndIf
  \If{$s<S$}
   \State Sample $\epsilon_{i,s}\sim\mathcal N(0,I)$; $X_i^{(s+1)}\gets\Psi_s(\widehat X_i^{(s)},\epsilon_{i,s})$
  \EndIf
 \EndFor
 \State $\mathcal H\gets\operatorname{Commit}_L(\mathcal H,\mathrm{KV}_i^\star)$; emit $\widehat X_i^{(S)}$
\EndFor
\end{algorithmic}
\end{algorithm}

\paragraph{Two-pass SDF training.}
Algorithm~\ref{alg:general_sdf} implements the two passes of Appendix~\ref{sec:sdf}; score-model updates follow the standard DMD procedure.
A context-stage plan $\{c_j\}\sim\pi(\cdot\mid d)$, drawn according to Eq.~\ref{eq:history_policies}, is shared by both passes.
In Pass~1, $\operatorname{SelectHistory}$ reads the preceding blocks at stage $k$ under the fixed policy, at the planned stages under the random policy, and at the current stage $s$ under the aligned policy, while $\operatorname{CommitPlan}_L$ retains only the K/V that the policy requires.
Both passes share the clean sink recached from $\widehat X_{\Omega}^{(d)}$, and the recorded trajectory $\mathcal R$ receives no gradients.

\begin{algorithm}[!t]
\caption{Two-Pass Training of SDF}
\label{alg:general_sdf}
\footnotesize
\begin{algorithmic}[1]
\Require Denoiser $f_\theta$, stages $t_{1:S}$, sampler $\Psi$, $N$ blocks, window $L$, sink $\Omega$, prompt $p$
\Require History policy $\pi$, dual-stream causal mask $M_{\mathrm{dual}}$
\Ensure One generator update
\State Sample $d\sim\mathcal{U}\{1,\ldots,S\}$ and context-stage plan $\{c_j\}\sim\pi(\cdot\mid d)$
\Statex \textbf{Pass 1: detached trajectory collection}
\State Disable gradients; roll out the sink blocks sequentially; record $X_{\Omega}^{(1:S)}$ and $\widehat X_{\Omega}^{(d)}$
\State Encode $\widehat X_{\Omega}^{(d)}$ at $t=0$ into $\mathrm{KV}_{\Omega}$ \Comment{sink recaching}
\State $\mathcal H\gets\operatorname{Init}(\mathrm{KV}_{\Omega})$; $\mathcal R_{i,s}\gets X_i^{(s)}$ for $i\in\Omega$ and all $s$
\For{$i=|\Omega|+1,\ldots,N$}
 \State Sample $X_i^{(1)}\sim\mathcal N(0,I)$
 \For{$s=1,\ldots,S$}
  \State $\mathcal R_{i,s}\gets\operatorname{sg}(X_i^{(s)})$
  \State $\mathcal H_i^{(s)}\gets\operatorname{SelectHistory}(\mathcal H,\pi,\{c_j\}_{j<i},s)$
  \State $(\widehat X_i^{(s)},\mathrm{KV}_i^{(s)})\gets f_\theta(X_i^{(s)},t_s;\mathcal H_i^{(s)},p)$
  \If{$s<S$}
   \State Sample $\epsilon_{i,s}\sim\mathcal N(0,I)$; $X_i^{(s+1)}\gets\Psi_s(\widehat X_i^{(s)},\epsilon_{i,s})$
  \EndIf
 \EndFor
 \State $\mathcal H\gets\operatorname{CommitPlan}_L(\mathcal H,\mathrm{KV}_i^{(1:S)},\pi,c_i)$
\EndFor
\State Discard the rollout cache $\mathcal H$
\Statex \textbf{Pass 2: differentiable replay}
\State $C_i\gets\mathcal R_{i,c_i}$ and $D_i\gets\mathcal R_{i,d}$ for $i=1,\ldots,N$
\State Pack $Z\gets[\widehat X_{\Omega}^{(d)}\mid C\mid D]$ and timestep vector $\mathbf t\gets[0\mid\{t_{c_i}\}\mid t_d]$; enable gradients
\State $\widehat X_D\gets[f_\theta(\operatorname{sg}(Z),\mathbf t;M_{\mathrm{dual}},p)]_D$ \Comment{clean sink recached with gradients}
\State $\mathcal L\gets\mathcal L_{\mathrm{DMD}}(\widehat X_D,p)$
\State Backpropagate through sink, context, and query K/V, but not into $\mathcal R$ or Pass 1; update $\theta$
\end{algorithmic}
\end{algorithm}

\section{Experimental Details}
\label{sec:evaluation_details}

\subsection{Implementation}
\label{sec:implementation_details}

\paragraph{Distillation setup.}
The generator is a causal Wan2.1-T2V-1.3B~\citep{wan2025wan} trained with DMD~\citep{yin2024onestep,yin2024improved}.
Wan2.1-T2V-14B serves as the real score model with a classifier-free guidance~\citep{ho2022classifier} scale of $3.0$, and the bidirectional Wan2.1-T2V-1.3B serves as the fake score model.
Following SGF~\citep{zhuang2026self}, we initialize SAF and SDF from the \texttt{ar\_diffusion} checkpoint released by Causal-Forcing~\citep{zhu2026causal}.
Training uses text prompts only, taken from the VidProM prompts~\citep{wang2024vidprom} as filtered and extended by Self-Forcing~\citep{huang2026self}.
Each training sample has 21 latent frames, corresponding to a 5-second video at $832\times480, 16$ FPS.
All models use $S=4$ denoising stages with noise levels $(t_1,\ldots,t_4)=(1000,750,500,250)$ and a timestep shift of $5$.

\paragraph{Attention.}
In the \emph{chunkwise} setting, each block contains 3 latent frames, and the local attention window spans 12 latent frames, including the current block and a 3-frame sink.
In the \emph{framewise} setting, each block contains 1 latent frame, and the window spans 21 latent frames with a 4-frame sink, following SGF~\citep{zhuang2026self}.
The window works as a first-in, first-out queue, except for the sink, which stays fixed.
The sink holds the latent frames at the start of the video and is recached at timestep zero from a single estimate, taken at the exit stage during training and at the final stage at inference, whereas the rest of the history keeps the noisy K/V from the denoising trajectory (Appendix~\ref{sec:algorithm_details}).

\paragraph{Optimization.}
We use AdamW~\citep{loshchilov2019decoupled} with $(\beta_1,\beta_2)=(0,0.999)$ and a weight decay of $0.01$.
The learning rate is $2\times10^{-6}$ for the generator and $4\times10^{-7}$ for the fake score model.
The fake score model is first warmed up for 10 updates, after which the generator is updated once every 5 fake score updates.
We train for 800 steps on 8 NVIDIA GB300 GPUs with a batch size of 4 per GPU (32 in total) and a random seed of $42$.
Starting at step 200, we also track an exponential moving average (EMA) of the generator weights with a decay of $0.99$.

\paragraph{Inference.}
We evaluate the EMA weights at step 600 for the chunkwise model and at step 700 for the framewise model.
SAF$^{\mathrm M}$ maintains a separate K/V bank for each denoising stage, whereas SAF$^{\mathrm S}$ maintains a single bank written by the Mix policy (Appendix~\ref{sec:algorithm_details}).
Inference uses the base seed $42$, from which per-sample seeds are derived as described in Appendix~\ref{sec:benchmark_details}.

\subsection{Benchmarks and Metrics}
\label{sec:benchmark_details}

Table~\ref{tab:benchmark_summary} summarizes the three benchmarks.
Seeding is deterministic in all evaluations.
The seed of each sample is derived from the base seed $42$, its global prompt index, and its sample index, so the results do not depend on how prompts are split across GPUs or on which subset is evaluated.

\begin{table}[!ht]
\centering
\caption{\textbf{Evaluation benchmarks.} All models are trained on 5-second clips, so Interactive and MovieGen-100s test generation at $12{\times}$ and $20{\times}$ the training length. Each Interactive video is driven by six prompts in turn, each lasting 10 seconds, and the benchmark provides 100 such prompt sequences ($100\times6$). Samples is the number of videos per prompt (per prompt sequence for Interactive), and Dimensions is the number of VBench dimensions scored.}
\label{tab:benchmark_summary}
\small
\setlength{\tabcolsep}{6pt}
\renewcommand{\arraystretch}{1.1}
\begin{tabular}{lccccc}
\toprule
Benchmark & Prompts & Samples & Length & Dimensions & Aggregate metrics \\
\midrule
VBench & 944 & 5 & 5\,s & 16 & Total, Quality, Semantic \\
Interactive & $100\times6$ & 1 & 60\,s & 7 & Quality, ViCLIP \\
MovieGen-100s & 128 & 1 & 100\,s & 7 & Quality \\
\bottomrule
\end{tabular}
\end{table}

\paragraph{Protocols.}
All three benchmarks follow existing evaluation protocols.
On \textbf{VBench}~\citep{huang2024vbench,huang2025vbenchpp}, we generate five videos for each of the 944 unique standard prompts and report the Total, Quality, and Semantic scores.
\textbf{Interactive} uses the 100 prompt sequences released by MemFlow~\citep{ji2025memflow} and the protocol of LongLive~\citep{yang2025longlive}.
Each 60-second video is generated from six prompts in turn, one for every 10 seconds.
Prompt alignment is the ViCLIP~\citep{wang2024internvid} similarity between each 10-second segment and its own prompt, averaged over the six segments.
\textbf{MovieGen-100s} uses the 128 extended MovieGen~\citep{polyak2025moviegen} prompts released by UniTemp~\citep{zhang2026unitemp} and generates one 100-second video per prompt.
On Interactive and MovieGen-100s, Quality is computed with VBench-Long~\citep{huang2025vbenchpp}, which splits each video into 2-second clips and scores them on the same seven quality dimensions as VBench.
All per-dimension scores in the tables are multiplied by $100$.

\subsection{Baselines and Throughput}
\label{sec:result_provenance}

We re-evaluate Self-Forcing~\citep{huang2026self}, LongLive~\citep{yang2025longlive}, Causal-Forcing~\citep{zhu2026causal}, HiAR~\citep{zou2026hiar}, and SGF~\citep{zhuang2026self} with their official weights and released inference settings, using the prompts and seeds described above.
Only Causal-Forcing and SGF release framewise checkpoints, so the framewise comparison (Table~\ref{tab:frame1_results}) includes only these two baselines.
Baseline throughput is taken from previously reported results.
For SAF, we measure throughput on H100 GPUs both on a single GPU and with 4-GPU pipelined inference (Algorithm~\ref{alg:pipeline_saf}), reported as the first and second values, respectively.

\section{Additional Quantitative Results}
\label{sec:full_results}

\paragraph{Framewise comparison.}
Table~\ref{tab:frame1_results} is the framewise counterpart of Table~\ref{tab:main_evaluation}.
SAF$^{\mathrm M}$ has the highest throughput, VBench Quality, and Interactive ViCLIP.
SGF leads in VBench Total and Semantic, as well as in Quality on Interactive and MovieGen-100s.
SAF$^{\mathrm M}$ outperforms SAF$^{\mathrm S}$ on all six aggregate scores, and the gap between the two inference policies is larger than in the chunkwise setting (e.g., $3.40$ versus $0.66$ in MovieGen-100s Quality).
We attribute this gap to error accumulation in single-bank inference, which is stronger in the framewise setting: each block contains a single latent frame that cannot be refined jointly with its neighbors, and the larger 21-frame window keeps more of this history in context (Appendix~\ref{sec:limitations}).

\begin{table}[!tbp]
\centering
\caption{\textbf{Framewise evaluation with existing methods.} Framewise counterpart of Table~\ref{tab:main_evaluation}; Causal-Forcing and SGF are the only baselines with released framewise checkpoints. Throughput is reported as in Table~\ref{tab:main_evaluation}, with the second value denoting 4-GPU pipelined inference. Best results are \textbf{bold} and second-best results are \underline{underlined}.}
\label{tab:frame1_results}
\setlength{\tabcolsep}{2pt}
\renewcommand{\arraystretch}{1.03}
\begin{adjustbox}{width=0.9\linewidth,center}
\begin{tabular}{lccccccc}
        \toprule
        \multirow{2}{*}{Method}
        & \multirow{2}{*}{\shortstack{Throughput\\(FPS) $\uparrow$}}
        & \multicolumn{3}{c}{VBench}
        & \multicolumn{2}{c}{Interactive}
        & \multicolumn{1}{c}{MovieGen 100s} \\
        \cmidrule(lr){3-5}
        \cmidrule(lr){6-7}
        \cmidrule(lr){8-8}
        & & Total $\uparrow$ & Quality $\uparrow$ & Semantic $\uparrow$ & ViCLIP $\uparrow$ & Quality $\uparrow$ & Quality $\uparrow$ \\
        \midrule
        Causal-Forcing & $8.9$ & $83.86$ & $85.23$ & $\underline{78.35}$ & $21.82$ & $81.42$ & $80.62$ \\
        SGF & $8.9$ & $\mathbf{84.10}$ & $\underline{85.26}$ & $\mathbf{79.48}$ & $\underline{24.85}$ & $\mathbf{84.83}$ & $\mathbf{84.71}$ \\
        \cmidrule(lr){1-8}
        \textbf{SAF}$^{\mathrm{S}}$ & $\underline{9.07}$ & $83.20$ & $84.52$ & $77.93$ & $22.85$ & $81.95$ & $80.77$ \\
        \textbf{SAF}$^{\mathrm{M}}$ & $\mathbf{9.08}$ / $\mathbf{21.6}$ & $\underline{83.88}$ & $\mathbf{85.31}$ & $78.15$ & $\mathbf{25.23}$ & $\underline{84.35}$ & $\underline{84.17}$ \\
        \bottomrule
    \end{tabular}
\end{adjustbox}
\end{table}

\paragraph{Per-dimension results.}
Tables~\ref{tab:appendix_vbench_all}--\ref{tab:appendix_moviegen100_all} break down the aggregate scores in Tables~\ref{tab:main_evaluation} and~\ref{tab:frame1_results}.
Consistency and motion generally trade off against each other.
Causal-Forcing has the highest dynamic degree in five of the six benchmark--setting pairs but the lowest background consistency in all six.
SGF shows the opposite pattern in the framewise setting, with the highest background consistency and the lowest dynamic degree on all three benchmarks.
\begin{itemize}\setlength{\itemsep}{1pt}\setlength{\parskip}{0pt}\setlength{\topsep}{2pt}
    \item \textbf{VBench} (Table~\ref{tab:appendix_vbench_all}). In the chunkwise setting, SAF$^{\mathrm M}$ is best in temporal flickering (tied with SAF$^{\mathrm S}$) and object class, and second in dynamic degree, subject consistency, and background consistency. Other methods are stronger in color, spatial relationship, and the style dimensions. In the framewise setting, SAF$^{\mathrm M}$ leads in motion smoothness, aesthetic quality, and object class, while SGF leads in subject consistency, background consistency, and imaging quality.
    \item \textbf{Interactive} (Table~\ref{tab:appendix_interactive_all}). In the chunkwise setting, SAF$^{\mathrm M}$ reaches a dynamic degree of $73.97$, well above the best baseline ($56.20$). From the second segment onward, its ViCLIP is second only to LongLive, whereas the ViCLIP of Self-Forcing declines after every prompt switch. In the framewise setting, SAF$^{\mathrm M}$ achieves the best ViCLIP in every segment after the first.
    \item \textbf{MovieGen-100s} (Table~\ref{tab:appendix_moviegen100_all}). In the chunkwise setting, SAF$^{\mathrm M}$ is best in subject consistency (tied with SAF$^{\mathrm S}$), background consistency, and aesthetic quality, and nearly matches the dynamic degree of Causal-Forcing ($65.61$ versus $65.69$). In the framewise setting, SAF$^{\mathrm M}$ leads in subject consistency and aesthetic quality. SAF$^{\mathrm S}$ keeps strong motion but has the lowest imaging quality, consistent with the drift discussed in Appendix~\ref{sec:limitations}.
\end{itemize}

\begin{table}[!tbp]
\centering
\caption{\textbf{VBench per-dimension results.} All 16 dimensions of the 5-second evaluation; higher is better. Chunkwise (left) and framewise (right) results are ranked separately; best and second-best entries are \textbf{bold} and \underline{underlined}.}
\label{tab:appendix_vbench_all}
\fontsize{7}{8.2}\selectfont
\setlength{\tabcolsep}{1.3pt}
\renewcommand{\arraystretch}{1.05}
\begin{adjustbox}{width=\linewidth}
\begin{tabular}{lccccccc@{\hspace{6pt}}cccc}
\toprule
\multirow{2}{*}{Metric} & \multicolumn{7}{c}{Chunkwise} & \multicolumn{4}{c}{Framewise} \\
\cmidrule(lr){2-8}\cmidrule(lr){9-12}
 & \shortstack{Self-\\Forcing} & LongLive & \shortstack{Causal-\\Forcing} & HiAR & SGF & \textbf{SAF}$^{\mathrm{S}}$ & \textbf{SAF}$^{\mathrm{M}}$ & \shortstack{Causal-\\Forcing} & SGF & \textbf{SAF}$^{\mathrm{S}}$ & \textbf{SAF}$^{\mathrm{M}}$ \\
\midrule
Subject Consistency & 95.09 & 96.98 & 95.48 & 96.35 & \textbf{97.14} & 96.25 & \underline{97.05} & 91.01 & \textbf{96.91} & 95.48 & \underline{96.28} \\
Background Consistency & 96.10 & \textbf{96.92} & 95.86 & 96.31 & 96.35 & 96.58 & \underline{96.62} & 92.85 & \textbf{96.02} & 94.84 & \underline{95.14} \\
Temporal Flickering & 99.01 & \underline{99.35} & 99.22 & 99.18 & 99.07 & \textbf{99.54} & \textbf{99.54} & \textbf{99.48} & \underline{99.11} & 98.66 & 98.71 \\
Motion Smoothness & 98.24 & \textbf{98.79} & 97.53 & 98.35 & \underline{98.51} & 98.46 & 98.39 & 97.25 & 98.23 & \underline{98.51} & \textbf{98.54} \\
Dynamic Degree & 66.38 & 40.83 & \textbf{84.44} & 47.22 & 66.94 & 71.39 & \underline{76.39} & \textbf{99.44} & 64.72 & 66.94 & \underline{72.78} \\
Aesthetic Quality & 65.79 & 67.03 & 66.63 & 66.51 & \textbf{68.32} & 67.68 & \underline{67.83} & 65.44 & \underline{67.21} & 66.83 & \textbf{67.30} \\
Imaging Quality & 69.71 & 69.18 & \underline{70.31} & 67.92 & \textbf{70.58} & 69.04 & 68.88 & 69.02 & \textbf{71.20} & 69.22 & \underline{69.38} \\
Object Class & 93.16 & 96.28 & 95.17 & 95.21 & 95.63 & \underline{96.49} & \textbf{96.61} & 93.89 & 95.52 & \underline{95.70} & \textbf{95.87} \\
Multiple Objects & \underline{87.19} & 86.49 & \textbf{87.82} & 82.93 & 86.83 & 85.27 & 84.82 & 75.63 & \textbf{84.91} & 81.43 & \underline{82.26} \\
Human Action & \underline{96.40} & 95.80 & 96.00 & \textbf{96.60} & \underline{96.40} & 95.20 & 96.00 & \textbf{96.00} & \underline{95.60} & 95.40 & \textbf{96.00} \\
Color & 86.83 & \textbf{90.79} & 86.16 & \underline{89.04} & 88.98 & 88.10 & 88.38 & \underline{83.92} & \textbf{87.75} & 83.29 & 83.79 \\
Spatial Relationship & \textbf{81.77} & 80.56 & \underline{80.98} & 77.33 & 80.32 & 77.67 & 78.22 & \textbf{74.49} & \underline{74.33} & 70.62 & 71.37 \\
Scene & \underline{56.13} & \textbf{58.79} & 55.00 & 53.85 & \underline{56.13} & 55.19 & 55.17 & \underline{55.36} & \textbf{55.74} & 54.68 & 54.42 \\
Appearance Style & 20.34 & 20.42 & \underline{20.53} & \textbf{20.56} & 20.46 & 20.21 & 20.20 & \textbf{20.72} & \underline{20.46} & 20.43 & 20.40 \\
Temporal Style & \underline{24.45} & 24.16 & \textbf{24.72} & 24.42 & 24.20 & 24.03 & 23.97 & \textbf{25.06} & \underline{23.90} & 23.73 & 23.57 \\
Overall Consistency & \textbf{26.85} & 26.61 & \underline{26.65} & \textbf{26.85} & 26.56 & 26.29 & 26.32 & \textbf{26.41} & \underline{26.26} & 26.12 & 26.12 \\
\bottomrule
\end{tabular}
\end{adjustbox}
\end{table}

\begin{table}[!tbp]
\centering
\caption{\textbf{Interactive per-dimension results.} The 7 VBench-Long quality dimensions (top) and ViCLIP prompt alignment for each 10-second segment (bottom); ViCLIP averages the six segments and equals the Interactive ViCLIP score in Tables~\ref{tab:main_evaluation} and~\ref{tab:frame1_results}. Chunkwise and framewise results are ranked separately; best and second-best entries are \textbf{bold} and \underline{underlined}.}
\label{tab:appendix_interactive_all}
\fontsize{8}{9.6}\selectfont
\setlength{\tabcolsep}{1.6pt}
\renewcommand{\arraystretch}{1.12}
\begin{adjustbox}{width=\linewidth}
\begin{tabular}{lccccccc@{\hspace{6pt}}cccc}
\toprule
\multirow{2}{*}{Metric} & \multicolumn{7}{c}{Chunkwise} & \multicolumn{4}{c}{Framewise} \\
\cmidrule(lr){2-8}\cmidrule(lr){9-12}
 & \shortstack{Self-\\Forcing} & LongLive & \shortstack{Causal-\\Forcing} & HiAR & SGF & \textbf{SAF}$^{\mathrm{S}}$ & \textbf{SAF}$^{\mathrm{M}}$ & \shortstack{Causal-\\Forcing} & SGF & \textbf{SAF}$^{\mathrm{S}}$ & \textbf{SAF}$^{\mathrm{M}}$ \\
\midrule
Subject Consistency & 97.38 & 97.74 & 95.14 & 97.69 & \underline{97.75} & \textbf{97.87} & 97.17 & 92.90 & \textbf{97.70} & 95.71 & \underline{97.05} \\
Background Consistency & 96.42 & \underline{96.58} & 95.08 & 96.12 & 96.55 & \textbf{96.70} & 96.17 & 94.14 & \textbf{96.64} & 95.16 & \underline{95.60} \\
Temporal Flickering & 99.07 & 99.04 & 98.47 & 98.34 & 98.68 & \underline{99.13} & \textbf{99.15} & 97.35 & \textbf{99.02} & 98.48 & \underline{98.58} \\
Motion Smoothness & \underline{99.10} & \textbf{99.16} & 97.68 & 98.66 & 98.97 & 99.00 & 98.83 & 97.77 & \textbf{98.86} & 97.61 & \underline{98.62} \\
Dynamic Degree & 22.73 & 27.47 & \underline{56.20} & 26.43 & 48.23 & 50.87 & \textbf{73.97} & \textbf{86.43} & 64.77 & \underline{75.13} & 69.67 \\
Aesthetic Quality & 57.77 & 60.82 & 57.30 & 57.74 & \textbf{61.65} & \underline{61.60} & 61.42 & 52.02 & \underline{60.00} & 54.72 & \textbf{60.32} \\
Imaging Quality & 68.04 & 71.45 & 69.00 & 71.54 & \underline{72.66} & \textbf{72.74} & 71.90 & 64.17 & \textbf{71.92} & 63.44 & \underline{70.24} \\
\midrule
ViCLIP & 21.88 & \textbf{25.74} & 24.21 & 24.16 & 24.63 & 24.70 & \underline{25.34} & 21.82 & \underline{24.85} & 22.85 & \textbf{25.23} \\
ViCLIP Segment 1 & 27.12 & \underline{27.48} & \textbf{27.51} & 27.25 & 27.14 & 27.10 & 27.06 & \textbf{27.53} & \underline{26.70} & 26.63 & 26.53 \\
ViCLIP Segment 2 & 24.54 & \textbf{26.32} & 25.57 & 23.46 & 24.62 & 24.88 & \underline{26.01} & 21.71 & \underline{25.10} & 22.97 & \textbf{25.25} \\
ViCLIP Segment 3 & 22.98 & \textbf{25.49} & 23.76 & 23.72 & 24.18 & 24.44 & \underline{25.24} & 19.44 & \underline{24.69} & 22.56 & \textbf{25.05} \\
ViCLIP Segment 4 & 20.11 & \textbf{24.95} & 23.38 & 23.17 & 23.55 & 23.67 & \underline{24.27} & 21.41 & \underline{23.88} & 22.04 & \textbf{24.37} \\
ViCLIP Segment 5 & 19.15 & \textbf{25.33} & 22.30 & 23.73 & 24.28 & 24.10 & \underline{24.78} & 18.95 & \underline{24.47} & 21.68 & \textbf{24.82} \\
ViCLIP Segment 6 & 17.38 & \textbf{24.89} & 22.74 & 23.64 & 24.00 & 24.00 & \underline{24.68} & 21.88 & \underline{24.25} & 21.19 & \textbf{25.35} \\
\bottomrule
\end{tabular}
\end{adjustbox}
\end{table}

\begin{table}[!tbp]
\centering
\caption{\textbf{MovieGen-100s per-dimension results.} The 7 VBench-Long quality dimensions over the full 100-second videos. Chunkwise and framewise results are ranked separately; best and second-best entries are \textbf{bold} and \underline{underlined}.}
\label{tab:appendix_moviegen100_all}
\fontsize{8}{9.6}\selectfont
\setlength{\tabcolsep}{1.6pt}
\renewcommand{\arraystretch}{1.12}
\begin{adjustbox}{width=\linewidth}
\begin{tabular}{lccccccc@{\hspace{6pt}}cccc}
\toprule
\multirow{2}{*}{Metric} & \multicolumn{7}{c}{Chunkwise} & \multicolumn{4}{c}{Framewise} \\
\cmidrule(lr){2-8}\cmidrule(lr){9-12}
 & \shortstack{Self-\\Forcing} & LongLive & \shortstack{Causal-\\Forcing} & HiAR & SGF & \textbf{SAF}$^{\mathrm{S}}$ & \textbf{SAF}$^{\mathrm{M}}$ & \shortstack{Causal-\\Forcing} & SGF & \textbf{SAF}$^{\mathrm{S}}$ & \textbf{SAF}$^{\mathrm{M}}$ \\
\midrule
Subject Consistency & 96.64 & 97.64 & 95.06 & 97.53 & \underline{98.02} & \textbf{98.06} & \textbf{98.06} & 92.98 & \underline{97.70} & 95.05 & \textbf{97.88} \\
Background Consistency & 96.34 & 96.65 & 95.49 & 96.23 & 96.81 & \underline{96.86} & \textbf{96.93} & 94.60 & \textbf{96.63} & 94.98 & \underline{96.49} \\
Temporal Flickering & 98.52 & \textbf{98.99} & 96.02 & 98.36 & 98.48 & 98.67 & \underline{98.73} & 96.28 & \textbf{98.70} & 97.70 & \underline{98.24} \\
Motion Smoothness & 98.24 & \textbf{98.68} & 95.83 & 98.36 & \underline{98.60} & 98.45 & 98.26 & 96.79 & \textbf{98.25} & 96.10 & \underline{98.00} \\
Dynamic Degree & 28.91 & 44.44 & \textbf{65.69} & 39.80 & 58.15 & 53.95 & \underline{65.61} & \textbf{85.16} & 67.10 & \underline{80.79} & 67.82 \\
Aesthetic Quality & 53.20 & 61.41 & 56.97 & 58.74 & 64.00 & \underline{64.03} & \textbf{64.59} & 52.50 & \underline{62.42} & 55.57 & \textbf{63.02} \\
Imaging Quality & 62.23 & 68.48 & 66.08 & 70.51 & \underline{71.66} & \textbf{71.80} & 70.09 & 64.70 & \textbf{70.55} & 60.35 & \underline{68.14} \\
\bottomrule
\end{tabular}
\end{adjustbox}
\end{table}

\section{Ablation Details}
\label{sec:ablation_details}

\subsection{Component and K/V Bank Ablations}

\paragraph{Setup.}
Tables~\ref{tab:inference_ablation} and~\ref{tab:appendix_inference_ablation} ablate two training components of SAF and the K/V bank used at inference, all in the chunkwise setting.
\emph{w/o K/V grad.} detaches the history K/V during training.
Every block is still supervised, but gradients from later blocks no longer flow into the K/V of earlier blocks.
\emph{w/o sink recache} skips the timestep-zero recache of the sink, so the sink keeps stage-aligned noisy K/V like the rest of the history.
Both training ablations are evaluated with the same SAF$^{\mathrm M}$ and SAF$^{\mathrm S}$ inference policies as the full model.
Fixed$t$ instead runs the full SAF checkpoint with a single bank written only at noise level $t$.
For each Interactive prompt sequence, we generate only the first 20 seconds, which cover the first two prompts, and score them with VBench-Long as in Appendix~\ref{sec:benchmark_details}.
ViCLIP is averaged over the two 10-second segments.

\paragraph{Results.}
Table~\ref{tab:appendix_inference_ablation} extends Table~\ref{tab:inference_ablation} to all seven quality dimensions and both ViCLIP segments.
Removing K/V gradients lowers dynamic degree under both inference policies, from $64.10$ to $51.30$ for SAF$^{\mathrm M}$ and from $48.00$ to $44.50$ for SAF$^{\mathrm S}$.
The slightly higher consistency and imaging quality come with this reduced motion.
Removing sink recaching leads to nearly static videos.
Dynamic degree falls below $7$, while subject consistency, background consistency, and motion smoothness reach their highest values within each group.
Second-segment ViCLIP also drops by more than $3$ points, showing that the model fails to follow the new prompt.
Under Fixed$t$, a larger $t$ monotonically increases Quality and dynamic degree but lowers motion smoothness and imaging quality.
The Mix policy of SAF$^{\mathrm S}$ falls between Fixed750 and Fixed500 in Quality and dynamic degree, and between Fixed500 and Fixed250 in motion smoothness and imaging quality.
It also achieves a higher ViCLIP score than any fixed bank ($25.99$ versus at most $25.71$).

\begin{table}[!tbp]
\centering
\caption{\textbf{Full component and K/V bank ablations on 20-second Interactive generation} (expanding Table~\ref{tab:inference_ablation}). Columns are grouped by inference policy: SAF$^{\mathrm{M}}$ (multiple stage-aligned banks), SAF$^{\mathrm{S}}$ (one Mix bank), and Fixed$t$ (one bank written at noise level $t$). In the first two groups, the w/o columns are training ablations evaluated with the same inference policy. ViCLIP averages the scores of the two 10-second segments. Best per group in \textbf{bold}, second \underline{underlined}.}
\label{tab:appendix_inference_ablation}
\fontsize{8.5}{10.1}\selectfont
\setlength{\tabcolsep}{4pt}
\renewcommand{\arraystretch}{1.12}
\begin{adjustbox}{width=\linewidth}
\begin{tabular}{lccc@{\hspace{8pt}}ccc@{\hspace{8pt}}ccc}
\toprule
\multirow{2}{*}{Metric} & \multicolumn{3}{c}{SAF$^{\mathrm{M}}$ (Aligned)} & \multicolumn{3}{c}{SAF$^{\mathrm{S}}$ (Mix)} & \multicolumn{3}{c}{Fixed$t$} \\
\cmidrule(lr){2-4}\cmidrule(lr){5-7}\cmidrule(lr){8-10}
 & Full & \shortstack{w/o K/V\\grad.} & \shortstack{w/o sink\\recache} & Full & \shortstack{w/o K/V\\grad.} & \shortstack{w/o sink\\recache} & $t{=}750$ & $t{=}500$ & $t{=}250$ \\
\midrule
Quality & \textbf{85.18} & \underline{84.64} & 81.70 & \textbf{84.22} & \underline{84.02} & 81.30 & \textbf{84.68} & \underline{84.12} & 83.99 \\
ViCLIP & \textbf{26.54} & \underline{26.15} & 24.42 & \underline{25.99} & \textbf{26.01} & 24.21 & \textbf{25.71} & 25.40 & \underline{25.65} \\
\midrule
Subject Consistency & 97.58 & \underline{98.07} & \textbf{99.24} & 98.04 & \underline{98.10} & \textbf{99.10} & \underline{98.04} & \textbf{98.25} & 97.99 \\
Background Consistency & 96.56 & \underline{97.01} & \textbf{98.04} & 96.92 & \underline{97.04} & \textbf{97.73} & \underline{96.90} & \textbf{97.00} & \underline{96.90} \\
Temporal Flickering & 99.26 & \textbf{99.38} & \underline{99.33} & \underline{99.28} & 99.23 & \textbf{99.29} & \textbf{99.24} & \underline{99.16} & 99.13 \\
Motion Smoothness & 98.93 & \underline{99.07} & \textbf{99.48} & 99.06 & \underline{99.07} & \textbf{99.41} & 98.94 & \underline{99.02} & \textbf{99.10} \\
Dynamic Degree & \textbf{64.10} & \underline{51.30} & 6.90 & \textbf{48.00} & \underline{44.50} & 4.30 & \textbf{55.60} & \underline{47.10} & 45.70 \\
Aesthetic Quality & 62.10 & \underline{62.23} & \textbf{62.58} & \underline{61.99} & 61.80 & \textbf{62.45} & \textbf{62.02} & \underline{61.99} & 61.83 \\
Imaging Quality & \underline{71.82} & \textbf{72.58} & 71.24 & \underline{72.23} & \textbf{72.72} & 71.04 & 71.93 & \underline{72.13} & \textbf{72.36} \\
\midrule
ViCLIP Segment 1 & \textbf{27.06} & \underline{26.69} & 26.59 & \textbf{27.10} & \underline{26.77} & 26.61 & 26.44 & \underline{26.49} & \textbf{27.14} \\
ViCLIP Segment 2 & \textbf{26.01} & \underline{25.61} & 22.25 & \underline{24.88} & \textbf{25.24} & 21.81 & \textbf{24.97} & \underline{24.31} & 24.16 \\
\bottomrule
\end{tabular}
\end{adjustbox}
\end{table}

\subsection{Training Schedule Ablations}
\label{sec:schedule_ablations}

\paragraph{Setup.}
Figure~\ref{fig:mode_ratio_quality} and Table~\ref{tab:appendix_mode_ratio} compare SDF training schedules $a/b/c$, which sample the random, fixed, and aligned policies of Eq.~\ref{eq:history_policies} with probabilities $a\%$, $b\%$, and $c\%$ (Appendix~\ref{sec:sdf}).
SAF is trained in a single pass (Algorithm~\ref{alg:aligned_saf}), and the other schedules use the two-pass SDF procedure (Algorithm~\ref{alg:general_sdf}).
All schedules share the same initialization and hyperparameters.
Each model generates a 20-second video for each of the 128 MovieGen-100s prompts, scored with VBench-Long as above.

\begin{table}[!htb]
\centering
\caption{\textbf{Full results across SDF training schedules on 20-second MovieGen-100s generation} (expanding Figure~\ref{fig:mode_ratio_quality}). Training schedules give the random/fixed/aligned history probabilities ($0/0/100$ is SAF), and each model is evaluated with Fixed250, Mix (SAF$^{\mathrm{S}}$), and Aligned (SAF$^{\mathrm{M}}$) inference. The columns after Quality are the 7 VBench-Long dimensions. Within each training schedule, best and second-best entries are \textbf{bold} and \underline{underlined}.}
\label{tab:appendix_mode_ratio}
\fontsize{8}{9.6}\selectfont
\setlength{\tabcolsep}{3.2pt}
\renewcommand{\arraystretch}{1.08}
\begin{adjustbox}{width=\linewidth}
\begin{tabular}{llcccccccc}
\toprule
Training & Inference & Quality & Subject & Background & Flicker & Smooth & Dynamic & Aesthetic & Imaging \\
\midrule
\multirow{3}{*}{\textbf{0/0/100} (SAF)} & Fixed250 & 83.97 & \textbf{98.12} & \textbf{96.87} & 98.64 & \textbf{98.59} & 50.39 & 63.98 & \underline{70.71} \\
 & Mix & \underline{84.59} & 98.01 & \underline{96.85} & \textbf{98.82} & \underline{98.47} & \underline{57.73} & \underline{64.27} & \textbf{70.87} \\
 & Aligned & \textbf{85.22} & \underline{98.06} & 96.82 & \underline{98.81} & 98.31 & \textbf{67.97} & \textbf{64.84} & 69.84 \\
\cmidrule(lr){1-10}
\multirow{3}{*}{0/100/0} & Fixed250 & \underline{83.34} & 97.11 & 96.20 & 97.43 & \textbf{98.76} & \underline{53.20} & 62.29 & \underline{71.70} \\
 & Mix & 82.57 & \textbf{98.15} & \textbf{96.99} & \underline{98.05} & \underline{98.68} & 31.48 & \underline{63.92} & \textbf{72.21} \\
 & Aligned & \textbf{84.81} & \underline{98.09} & \underline{96.95} & \textbf{98.29} & 98.26 & \textbf{62.27} & \textbf{64.76} & 71.41 \\
\cmidrule(lr){1-10}
\multirow{3}{*}{100/0/0} & Fixed250 & 83.40 & 98.34 & 97.12 & 98.67 & \textbf{98.99} & 36.09 & 64.81 & 71.29 \\
 & Mix & \underline{83.73} & \underline{98.36} & \underline{97.15} & \underline{98.72} & \underline{98.94} & \underline{39.14} & \underline{65.15} & \textbf{71.57} \\
 & Aligned & \textbf{84.67} & \textbf{98.50} & \textbf{97.37} & \textbf{98.78} & 98.84 & \textbf{50.08} & \textbf{65.64} & \underline{71.40} \\
\cmidrule(lr){1-10}
\multirow{3}{*}{60/20/20} & Fixed250 & 82.84 & \textbf{98.85} & \textbf{97.73} & \textbf{99.30} & \textbf{99.32} & 18.05 & \underline{65.61} & \underline{71.60} \\
 & Mix & \underline{82.99} & \textbf{98.85} & \underline{97.69} & \underline{99.24} & \underline{99.26} & \underline{20.39} & \textbf{65.73} & \textbf{71.71} \\
 & Aligned & \textbf{83.97} & 98.55 & 97.35 & 98.89 & 98.92 & \textbf{40.55} & 65.51 & 71.15 \\
\cmidrule(lr){1-10}
\multirow{3}{*}{30/35/35} & Fixed250 & 83.50 & \textbf{98.61} & \textbf{97.38} & \textbf{99.09} & \textbf{99.11} & 30.39 & \textbf{65.31} & \underline{72.02} \\
 & Mix & \underline{83.53} & \underline{98.57} & \underline{97.35} & \underline{99.01} & \underline{99.02} & \underline{31.95} & \underline{65.24} & \textbf{72.18} \\
 & Aligned & \textbf{84.74} & 98.30 & 97.02 & 98.61 & 98.53 & \textbf{55.86} & 65.11 & 71.75 \\
\bottomrule
\end{tabular}
\end{adjustbox}
\end{table}

\paragraph{Results.}
Table~\ref{tab:appendix_mode_ratio} reports all dimensions behind Figure~\ref{fig:mode_ratio_quality}.
Under every training schedule, Aligned inference gives the highest Quality and dynamic degree.
SAF also achieves the highest Quality under every inference policy ($83.97$, $84.59$, and $85.22$ for Fixed250, Mix, and Aligned), and with Aligned inference it achieves the best Quality and dynamic degree overall ($85.22$ and $67.97$), although its imaging quality is the lowest in the table ($69.84$).
Mix inference outperforms Fixed250 in Quality under every schedule except the fixed-only $0/100/0$.
This model always reads history at $t{=}250$ during training, so Fixed250 is the inference policy most consistent with its training, whereas Mix also sharply reduces its dynamic degree, from $53.20$ to $31.48$.
With Fixed250 or Mix inference, the $60/20/20$ schedule has the highest subject consistency, background consistency, temporal flickering, and motion smoothness, but the lowest dynamic degree.
This again reflects the consistency--motion trade-off noted in Appendix~\ref{sec:full_results}.

\section{Additional Qualitative Results}
\label{sec:qualitative_gallery}
\begingroup
\setlength{\intextsep}{6pt plus 2pt minus 2pt}
\setlength{\abovecaptionskip}{4pt}
\newcommand{\galleryw}{0.86\linewidth}

We show additional comparisons on Interactive and MovieGen-100s.
The chunkwise figures (Figures~\ref{fig:extra_frame3_interactive}--\ref{fig:movie100_0818_full}) compare SAF$^{\mathrm S}$ and SAF$^{\mathrm M}$ with all five baselines.
The framewise figures (Figures~\ref{fig:extra_frame1_interactive}--\ref{fig:extra_frame1_movie100_second}) compare them with Causal-Forcing and SGF, the two baselines that release framewise checkpoints.
Within each figure, all methods use the same prompt or prompt sequence, the same sample seed, and the inference settings of the quantitative evaluation.
Method names are shown on the left of each row.
Interactive figures show the middle frame from each 10-second prompt interval and list the six prompts below the grid.
MovieGen-100s figures show frames at 0, 20, 40, 60, 80, and 100 seconds.
More visualizations are provided in the supplementary material.

\section{Limitations}
\label{sec:limitations}

\paragraph{Multiple versus single K/V banks.}
On a single GPU, SAF$^{\mathrm M}$ needs more memory than SAF$^{\mathrm S}$ because it stores one K/V bank per denoising stage.
With pipelined inference (Algorithm~\ref{alg:pipeline_saf}), however, each GPU holds only the bank of its own stage, so the per-GPU memory is close to that of SAF$^{\mathrm S}$.
SAF$^{\mathrm S}$ saves memory by letting all stages read the same bank, which departs from the stage-aligned history used in training and allows errors to accumulate over long rollouts.
The effect is minor in the chunkwise setting but more visible in the framewise setting, where single-frame blocks and a 21-frame window keep more of this history in context (Tables~\ref{tab:main_evaluation} and~\ref{tab:frame1_results}).
We therefore recommend SAF$^{\mathrm M}$ with multi-GPU pipelined inference. SAF$^{\mathrm M}$ can also run on a single H100 GPU, so SAF$^{\mathrm S}$ is only needed when GPU memory is limited.

\paragraph{Inconsistent visual changes.}
We still observe inconsistent changes in two situations.
When the camera moves, content newly entering at the frame boundary can change abruptly across frames, because the model can only refer to noisy K/V in the local window, and nothing outside the window except the sink is kept.
At prompt switches, since SAF is trained only on single-prompt clips, changing the action is handled smoothly, but introducing new objects, scenes, or characters can make the appearance unstable, as the sink contains none of this content.
Selectively retaining informative K/V beyond the window and training with prompt switches may address these two cases, which we leave to future work.

\begin{figure}[H]
\centering
\includegraphics[width=\galleryw]{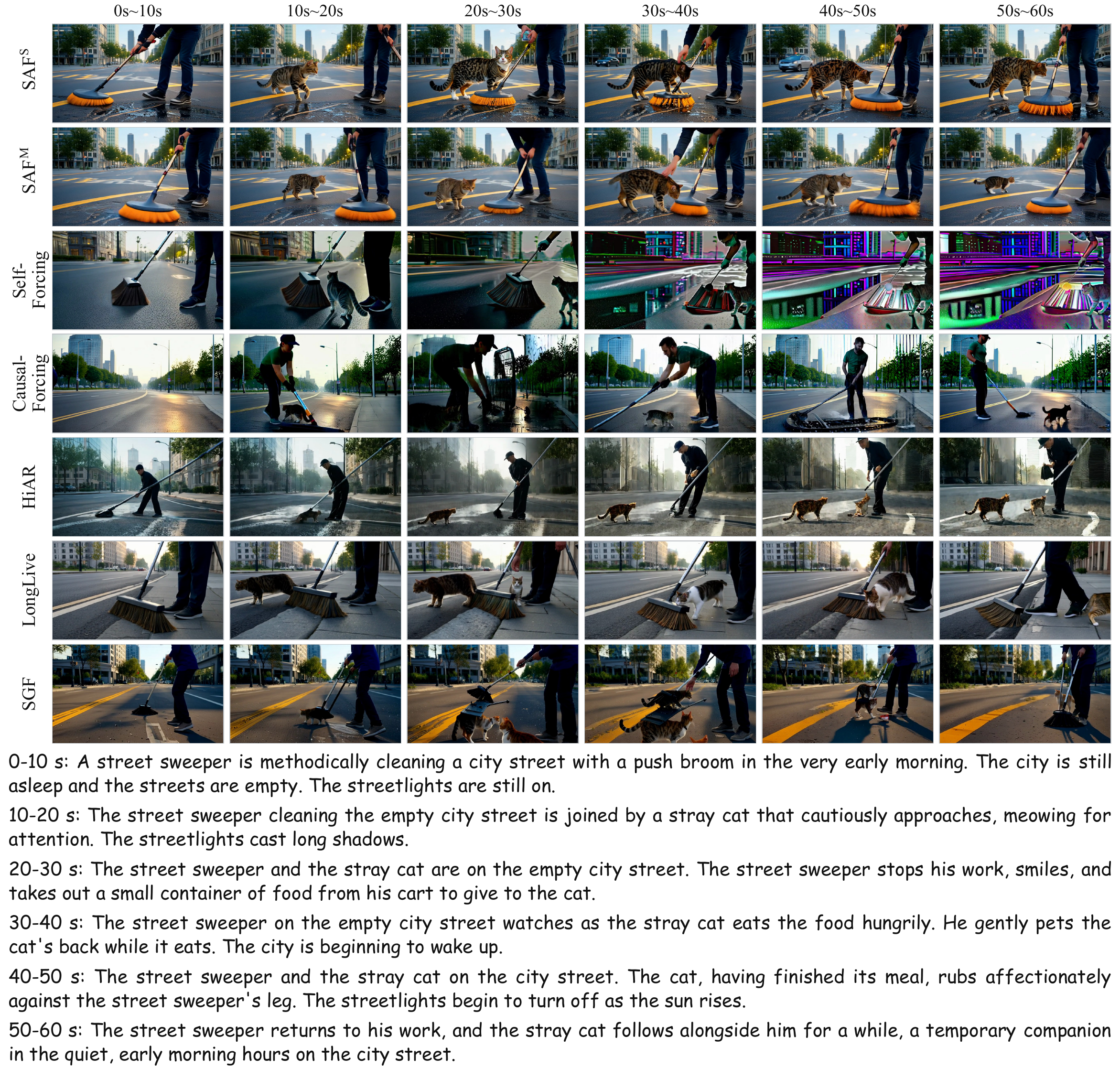}
\caption{\textbf{Additional chunkwise Interactive comparison.} Zoom in for better visualization.}
\label{fig:extra_frame3_interactive}
\end{figure}

\begin{figure}[H]
\centering
\includegraphics[width=\galleryw]{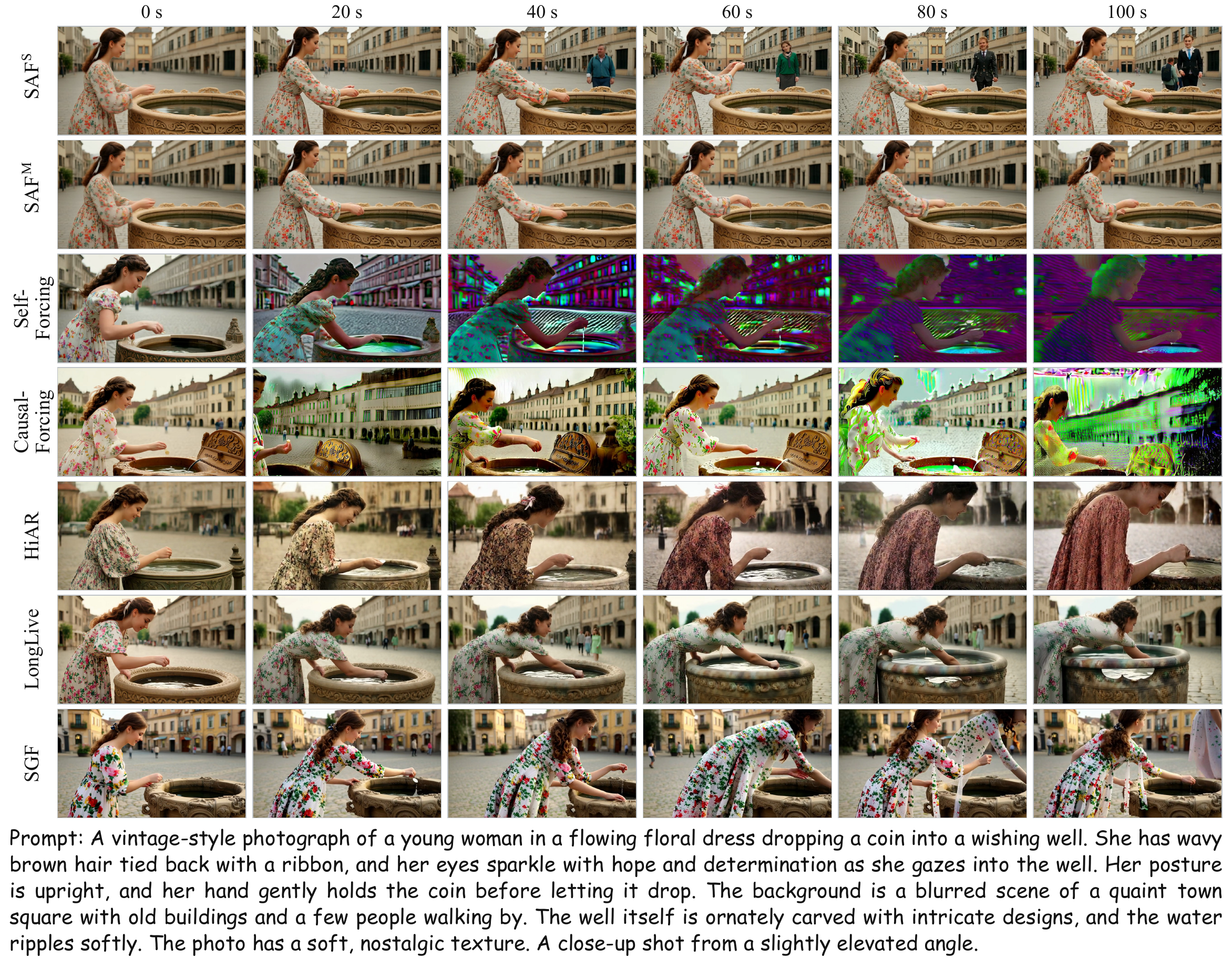}
\caption{\textbf{Additional chunkwise MovieGen-100s example 1.} Zoom in for better visualization.}
\label{fig:movie100_0483_full}
\end{figure}
\begin{figure}[H]
\centering
\includegraphics[width=\galleryw]{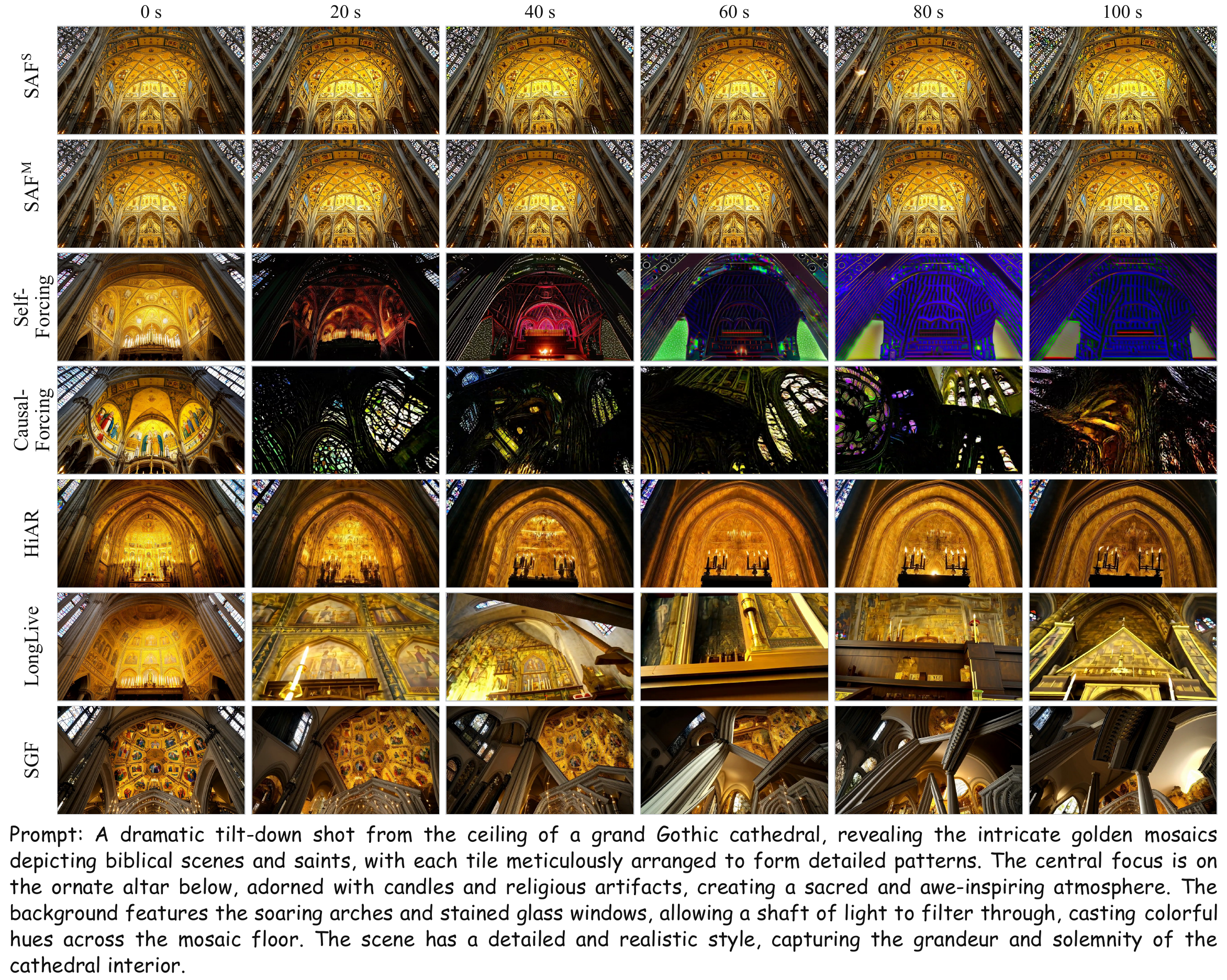}
\caption{\textbf{Additional chunkwise MovieGen-100s example 2.} Zoom in for better visualization.}
\label{fig:movie100_0818_full}
\end{figure}

\begin{figure}[H]
\centering
\includegraphics[width=\galleryw]{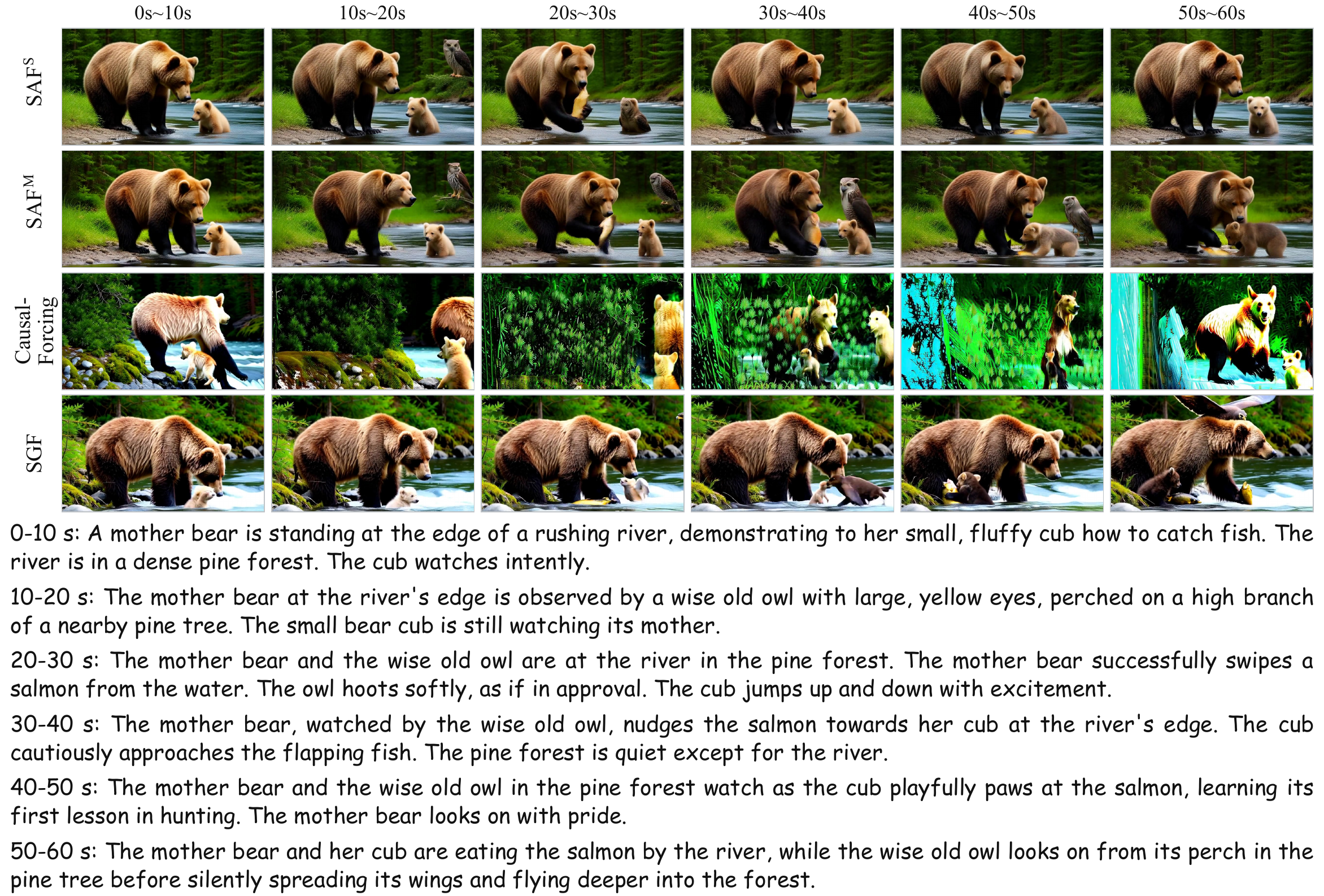}
\caption{\textbf{Additional framewise Interactive comparison.} Zoom in for better visualization.}
\label{fig:extra_frame1_interactive}
\end{figure}
\begin{figure}[H]
\centering
\includegraphics[width=\galleryw]{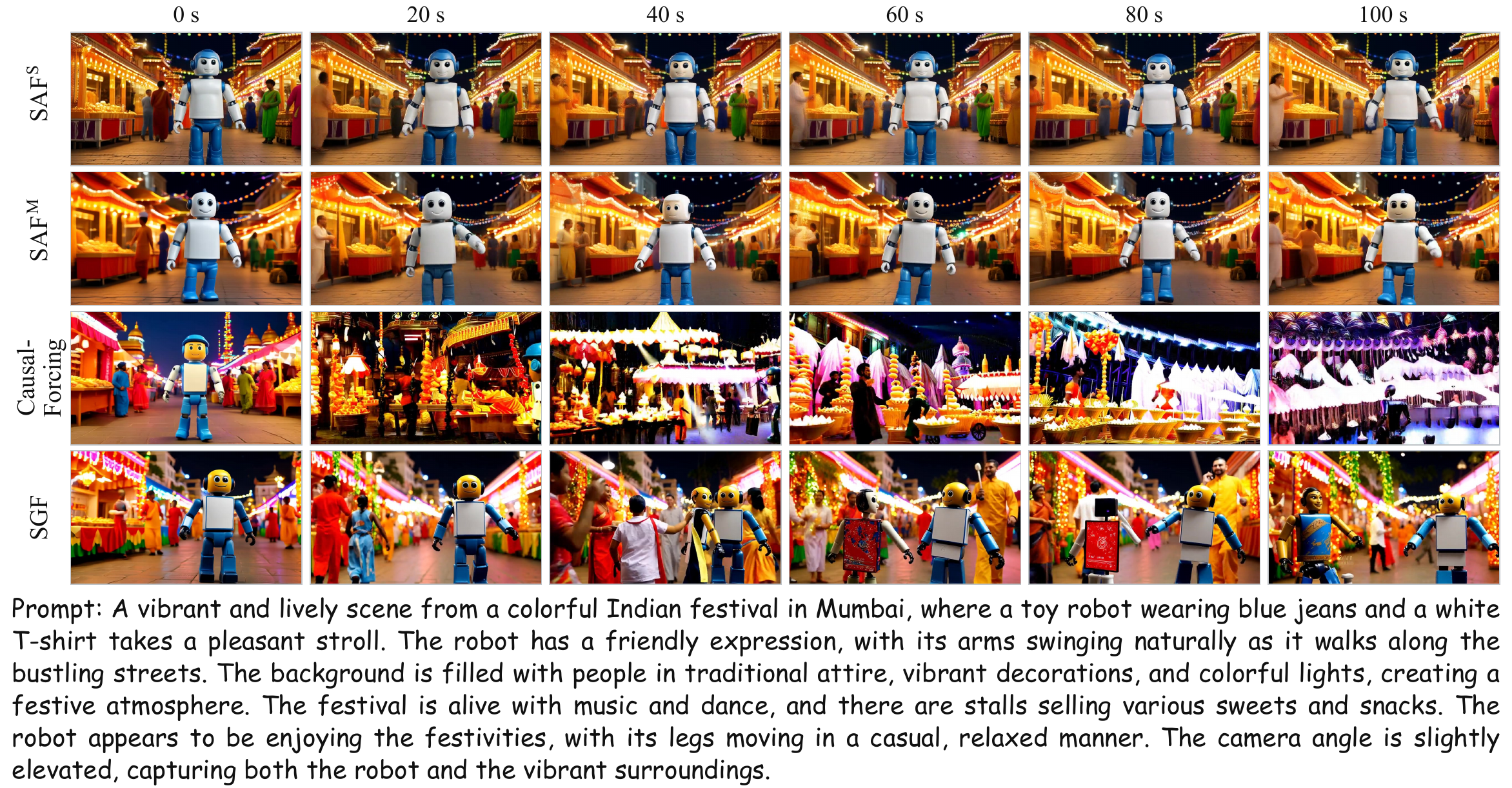}
\caption{\textbf{Additional framewise MovieGen-100s example 1.} Zoom in for better visualization.}
\label{fig:extra_frame1_movie100}
\end{figure}
\begin{figure}[H]
\centering
\includegraphics[width=\galleryw]{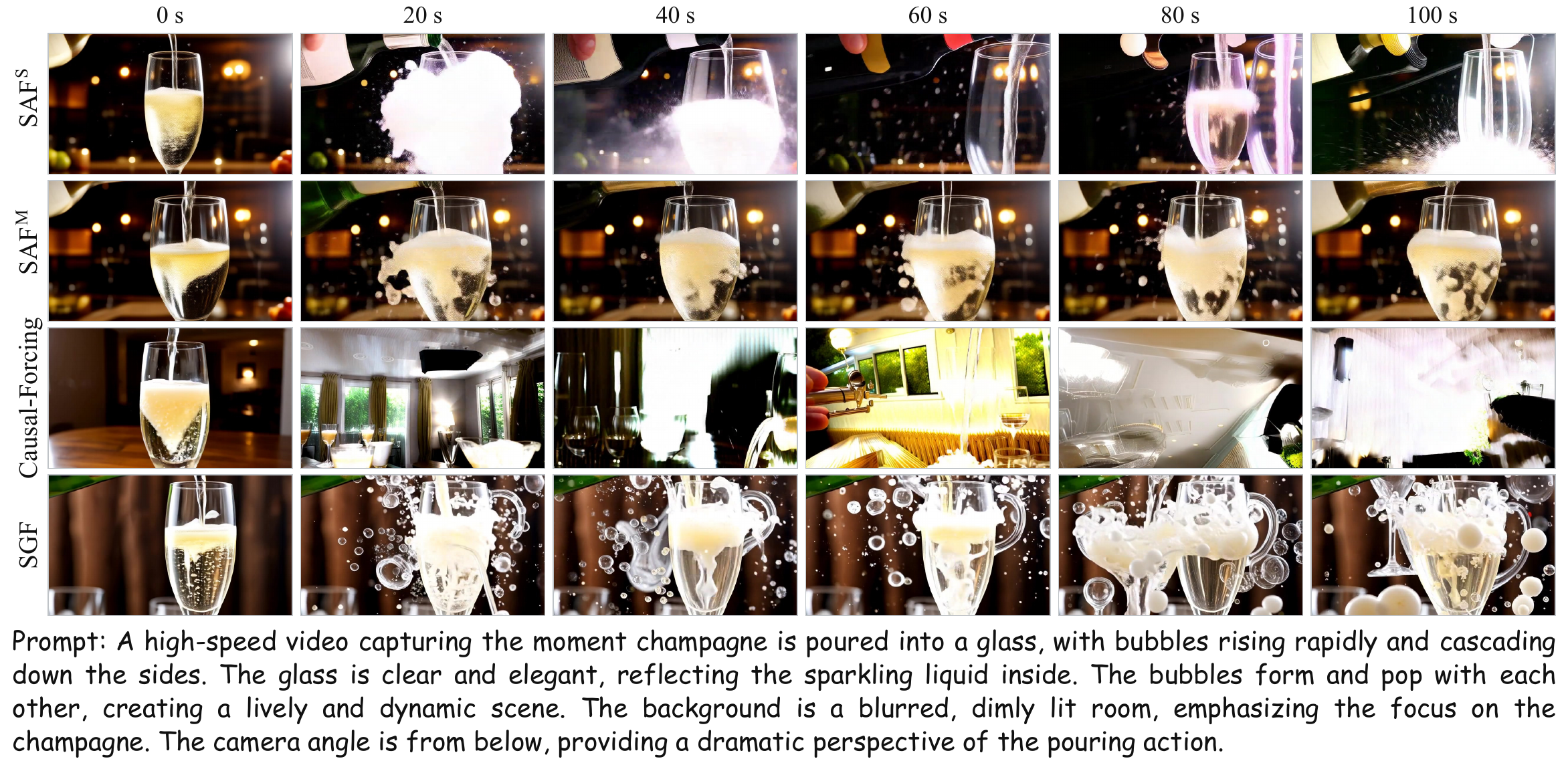}
\caption{\textbf{Additional framewise MovieGen-100s example 2.} Zoom in for better visualization.}
\label{fig:extra_frame1_movie100_second}
\end{figure}

\endgroup
\end{document}